\documentclass[11pt,letterpaper]{article}

\usepackage{orcidlink}

\usepackage{lmodern}
\usepackage{latexsym,multirow}
\usepackage{amssymb,amsmath,bm}
\usepackage{bigints}
\usepackage{array}
\usepackage{longtable}
\usepackage{booktabs}       
\usepackage{bbm} 
\usepackage{graphicx,float}
\usepackage[color,all,import,arrow]{xy}
\usepackage{enumerate}
\usepackage{verbatim}
\usepackage{lscape}
\usepackage{mathtools}
\usepackage{fancyvrb}
\usepackage{pdflscape} 

\usepackage[format=plain,
            labelfont={bf,it},
            textfont=it]{caption}

\usepackage{longtable}

\usepackage{placeins}

\usepackage[
  natbib=true,
  maxcitenames=1,
  maxbibnames=1,
  backend=biber,
  style=ext-authoryear
]{biblatex}
\usepackage{dcolumn}
\newcolumntype{.}{D{.}{.}{-1}}
\newcolumntype{d}[1]{D{.}{.}{#1}}

\usepackage{caption}
\usepackage{subcaption}
\DeclareCaptionLabelFormat{step}{Step \textbf{#2)}} 

\usepackage[T1]{fontenc}
\usepackage{tikz}
\usepackage{qtree}
\usepackage{comment}
\usetikzlibrary{arrows}
\usetikzlibrary{arrows.meta}
\usetikzlibrary{positioning}
\usetikzlibrary{matrix}
\usetikzlibrary{bayesnet}
\usetikzlibrary{decorations.pathreplacing,calligraphy} 
\usetikzlibrary{shapes.geometric}
\usetikzlibrary{calc}

\usepackage[ruled,vlined]{algorithm2e}

\usepackage{theorem}
\theoremstyle{plain}
\theoremheaderfont{\scshape}

\usepackage{chngpage}

\usepackage{rotating}

\usepackage{arydshln}

\usepackage[compact]{titlesec}

\usepackage{setspace}
\usepackage[bottom]{footmisc}
\allowdisplaybreaks

\newcommand\spacingset[1]{\renewcommand{\baselinestretch}%
{#1}\small\normalsize}

\newcommand{\blind}{0}

\newcommand{\bM}{\mathbf{M}}
\newcommand{\bmm}{\mathbf{m}}

\usepackage{xr}

\hypersetup{
    bookmarksopen=true,
    bookmarksnumbered=true, 
    pdfstartview=FitH,
    breaklinks=true,
    urlbordercolor={0 1 0},
    citebordercolor={0 0 1}
}

\usepackage{colortbl}

\begin{document} 

\newcommand{\tit}{
A Scoping Review of Earth Observation and Machine Learning for Causal Inference: Implications for the Geography of Poverty 
}

\spacingset{1.25}

\if0\blind

\title{\bf\tit\thanks{We thank Ola Hall and Ibrahim Wahab, and an anonymous reviewer for helpful comments. Thanks also go to the members of the AI \& Global Development Lab for continuing discussions and inspiration. See \href{https://GitHub.com/AIandGlobalDevelopmentLab/eo-poverty-review}{\url{GitHub.com/AIandGlobalDevelopmentLab/eo-poverty-review}} for a full listing of papers found in the systematic literature review. See \href{https://doi.org/10.7910/DVN/GN8CDJ}{\url{doi.org/10.7910/DVN/GN8CDJ}} for Supplementary Materials. 
}
  \author{    
      Kazuki Sakamoto
\thanks{PhD Candidate, Institute for Analytical Sociology, Link\"{o}ping University. Email:
      \href{mailto:kazuki.sakamoto@liu.se}{kazuki.sakamoto@liu.se}; ORCID: 0000-0003-0129-3338.
      }
            \and 
      Connor T. Jerzak
      \thanks{Assistant Professor, Department of
      Government, University of Texas at Austin. Email: \href{mailto:connor.jerzak@austin.utexas.edu}{connor.jerzak@austin.utexas.edu}; URL:
      \href{https://connorjerzak.com}{ConnorJerzak.com}; ORCID: 0000-0003-1914-8905.
      }
      \and 
      Adel Daoud
      \thanks{Senior Associate Professor, Institute for Analytical Sociology, Link\"{o}ping University;  Email: \href{mailto:adel.daoud@liu.se}{adel.daoud@liu.se}; URL:
      \href{http://adeldaoud.com/}{AdelDaoud.com}; ORCID: 0000-0001-7478-8345.
      }
  \date{
    \today
  }
}
}

\fi

\if1\blind
\title{\bf \tit}
\fi

\maketitle

\pdfbookmark[1]{Title Page}{Title Page}

\thispagestyle{empty}
\setcounter{page}{0}
         
\begin{abstract}
  \noindent Earth observation (EO) data such as satellite imagery can have far-reaching impacts on our understanding of the geography of poverty, especially when coupled with machine learning (ML) and computer vision. Early research used computer vision to predict living conditions in areas with limited data, but recent studies increasingly focus on causal analysis. Despite this shift, the use of EO-ML methods for causal inference lacks thorough documentation, and best practices are still developing. Through a comprehensive scoping review, we catalog the current literature on EO-ML methods in causal analysis. We synthesize five principal approaches to incorporating EO data in causal workflows:  (1) outcome imputation for downstream causal analysis, (2) EO image deconfounding, (3) EO-based treatment effect heterogeneity, (4) EO-based transportability analysis, and (5) image-informed causal discovery. Building on these findings, we provide a detailed protocol guiding researchers in integrating EO data into causal analysis—covering data requirements, computer vision model selection, and evaluation metrics. While our focus centers on health and living conditions outcomes, our protocol is adaptable to other sustainable development domains utilizing EO data. 
\vspace{.25cm} 
\vspace{0.1cm}
\end{abstract}


\clearpage
\spacingset{1}


\section{Introduction}

A shift is occurring in the geography of poverty research, driven by convergent influences from machine learning (ML), causal inference, and remote sensing disciplines. These emerging methods promise new insights into the processes underlying poverty---critical for meeting the United Nations’ first Sustainable Development Goal to eradicate poverty. Although development assistance rose to \$223.7 billion in 2023 and the global extreme poverty rate declined to 8.4\% in 2019 (down from 10.8\% in 2015), approximately 575 million people may still live in extreme poverty by 2030 \citep{oecd2024development}. Persistent poverty underscores the urgency to leverage new EO-ML methods to better fight poverty and provide more effective interventions that current methods may have missed.

On the one hand, research has long examined the geographic dimension of poverty, addressing a range of phenomena such as spatial poverty traps \citep{bird2010spatial, jalan1997spatial}, crime \citep{hipp2016general}, sustainable development \citep{briggs2018poor}, inequality \citep{chetty2014land}, and urban growth \citep{glaeser2011cities, lobao2007sociology}. One the other, recent technological advances in ML, defined as the ability of computers to learn from data without being explicitly programmed, expand upon these lines of inquiry by offering flexible modeling capabilities that capture more nuanced geo-temporal patterns than is feasible with traditional statistical methods. In parallel, ML has contributed to progress across diverse domains, including physics \citep{karniadakis2021physics}, Earth systems \citep{burke2021using}, health \citep{kino2021scoping}, and economics \citep{athey2018impact}, among others \citep{daoud_statistical_2023}. Within ML, computer vision techniques vision transformers are uniquely suited for analyzing satellite imagery and other remote sensing data. In this chapter, we use the term Earth Observation (EO) to encompass the systematic study of Earth’s physical, chemical, and biological systems via satellites, drones, or ground-based sensors \citep{runge_causal_2023, perez-suay_causal_2019}. When ML methods are applied to these forms of remotely sensed data, we refer to them collectively as EO-ML. Interest in leveraging EO-ML methods for poverty-related applications has notably increased, given its potential to provide fine-grained evidence about living conditions and resource needs \citep{hall2023review, jean2016combining, babenko2017poverty}.

In conjunction with these measurement-focused innovations, another stream of poverty research investigates the causal pathways linking poverty to health \citep{ridley2020poverty}, climate vulnerability \citep{ribot2017cause}, and broader developmental processes \citep{beck2007finance}. Addressing these pathways requires rigorous causal frameworks \citep{pearl2009causality} and further operationalized through improved research designs \citep{imbens2015causal}. Despite significant advances in the predictive use of EO-ML, its potential to strengthen causal inference in poverty research has only recently begun to emerge \citep{ratledge_using_2022,jerzak_causalimages_2023}. These early efforts suggest that combining EO-ML methods with a causal inference approach could improve our understanding of spatial and contextual factors influencing how poverty arises---and persists----in specific geographic settings. However, further investigations are required to determine under which conditions EO-ML methods are most effective in reliably uncovering causal relationships and clarifying their limitations when informing targeted interventions aimed at understanding geographic disparities. This task---examining these methodological strengths and constraints---is the central focus of this chapter.

To contextualize our work, Section \ref{s:ClassicalStat} discusses spatial statistical poverty literature that has established the groundwork for studying the geography of poverty. In Section \ref{s:ML}, we discuss advances in poverty mapping using outcome prediction from EO-ML models, where researchers have combined living condition maps with supplemental data such as mobile phone call record data \citep{blumenstock2015}, nighttime lights \citep{jean2016combining}, and social media data \citep{llorente2015social}. Our main contribution in Section \ref{s:Scoping} has two aims: first, to conduct a scoping review on recent literature around EO-ML methods for causal inference and secondly, to analyze our findings, to understand the current state of the art for the implications of the geography of poverty and identifying trends and research prospects. In Section \ref{s:protocol}, we present a detailed protocol for causal EO-ML analysis to guide future research. Although our review focuses on the geography of poverty, our protocol is potentially relevant to epidemiology, inequality, and other research on health and living conditions. Lastly, in Section \ref{s:FutureDirection}, we summarize our findings and discuss some future considerations for the geography of poverty researchers looking to include causal EO-ML methods into their research.

\section{Spatial Statistical Approaches to Poverty Prediction}\label{s:ClassicalStat}

The study of poverty is complex and intertwined across various disciplines: economics, sociology, public health, geography, and statistics. The geography of poverty represents a crucial component of this broader field, incorporating the spatial dimension. While this section is not a complete history of the geography of poverty, it provides context that underpins our subsequent recommendations for integrating causal EO-ML methodologies into the field.

\subsection{A Spatial Understanding of Poverty}

Poverty remains a difficult challenge throughout the world. Researchers have long explored the strong geographic dimensions of poverty, though a consensus on the precise nature of this relationship remains elusive. 

Although multidimensional measurements  \citep{alkire2011counting} of poverty are being developed and collected, monetary indicators such as asset and income remain dominant in the field \citep{powell2001towards}. Both approaches were influenced by the theoretical development spurred by \cite{sen1990development}'s work, specifically the capability approach and entitlement approach \citep{Reddy_Daoud_2020}. Not only did Sen's theory propose a new way of understanding the causes of poverty, but also how it should be measured and defined \citep{Daoud_2017Malthus}.  Nonetheless, physical, economic, and human capital indicators have enabled poverty researchers to scrutinize phenomena like poverty traps \citep{jalan1997spatial}. While spatial poverty traps are highly localized, they have been observed in numerous countries worldwide and are not unique to specific regions or nations. Theorists have posited their formation as due to deficits in geographical capital \citep{bird2019addressing}, fragile ecological environments \citep{yang2023ecological}, and political disadvantages \citep{burke2008spatial}. Some scholars emphasize strong geographic drivers tied to agricultural productivity, climate, and the spread of diseases \citep{sachs2001geography}, while others look to institutional factors \citep{acemoglu2010africa} as key drivers. 

\subsection{Small Area Estimation for Poverty Mapping}

Early poverty research often centered on national or state-level welfare data and grappled with how to set a rigorous poverty threshold \citep{orshansky1965counting}. Such large-area surveys, however, rarely allow for precise local targeting. Even contemporary data remain too sparse to provide small-area estimates across diverse geographies \citep{molina2014small}.

To address this gap, small area estimation models---like Fay-Herriot \citep{fay1979estimates} or Elbers, Lanjouw and Lanjouw \citep{elbers2003micro}---use indirect measures from census or other auxiliary data to project poverty rates at finer resolutions. This yields detailed poverty maps that can uncover local disparities concealed by national or regional averages \citep{bedi2007more,milbourne2010geographies}. Improvements in mapping technology during the 1990s further enhanced the accuracy and efficiency of such estimates, paving the way for machine learning to generate historical poverty maps and track inequality trends \citep{suss2024geowealth}.

\subsection{Spatial Econometrics}\label{s:spatialeconometrics}

Building on the recognition that poverty exhibits distinct spatial patterns, researchers developed econometric models to explicitly address these geographic dependencies. A foundational concept in this domain is \textit{spatial autocorrelation}, which quantifies the extent to which a variable’s values, such as poverty levels, are correlated across neighboring locations. This idea was first formalized by Cliff and Ord in their influential work from the late 1960s \citep{cliff1970spatial}, with its historical significance later emphasized by Getis \citep{getis2008history}. Unlike traditional correlation, which examines relationships between distinct variables, spatial autocorrelation focuses on a single variable’s spatial relationships across georeferenced units. When residual errors in standard regression models display spatial autocorrelation, it signals potential misspecification, often due to the omission of spatially structured explanatory variables \citep{li2024mitigating}. 

For example, if we sought to predict poverty levels in village $i$ in year $t$, a traditional regression model might use poverty levels in year $t-1$. But if we incorporate spatial autoregression, poverty levels in neighboring village $j$ also influence the poverty prediction in village $i$ \citep{wilson2014geographic}. The spatial lag model directly accounts for spatial autocorrelation; this would equate to neighboring poverty levels being included in the regression for the poverty of a specific village. 

To model such dependencies, researchers employ a spatial weights matrix, denoted \(\mathbf{W}\), an \(n \times n\) matrix where each entry \(w_{ij}\) defines the spatial relationship between units \(i\) and \(j\)—typically set with \(w_{ii} = 0\)—and can be constructed using criteria like contiguity, distance, or nearest neighbors \citep{griffith2020some}. In a spatial lag model, the outcome for a given unit \(i\), \(Y_i\), is influenced not only by its own attributes but also by the outcomes of neighboring units, expressed as \(\sum_{j=1}^{n} w_{ij} Y_j\). This model can be formalized as: 
\[
Y_i = \alpha + \rho \sum_{j=1}^{n} W_{ij} Y_j + \epsilon_i, \quad i = 1, \ldots, n,
\]
where 
\begin{itemize}
    \item \(W_{ij}\) is the \(ij\)-th element of \(\mathbf{W}\); 
    \item \(\sum_{j=1}^{n} W_{ij} Y_j\) is the spatially weighted average of the dependent variable; 
    \item \(\rho\) is the spatial autoregressive parameter that measures the intensity of the spatial interdependence (with \(\rho > 0\) indicating positive, \(\rho < 0\) indicating negative, and \(\rho = 0\) indicating no spatial autocorrelation); 
    \item \(\epsilon_i\) is the error term representing unobserved factors affecting the outcome. 
\end{itemize}
In other words, each contribution of ${W}_{ij}$ values determines how much the neighboring outcome $j$ values contribute to the outcome in the current location, $i$. Because this framework accounts for spatial dynamics in geographical poverty analysis, it was widely adopted in research aiming to quantify associations between various proposed causal factors and spatially defined poverty outcomes \citep{petrucci2003application}.\footnote{While spatial regression remains a dominant form of statistical analysis with geographic data, we note the emergence of some more recent methodologies that analyze the effect of spatial treatments using stochastic interventions \citep{papadogeorgou2022causal}.}

Building on the foundation of spatial autocorrelation and econometric models like the spatial lag framework just introduced, the field of spatial econometrics has undergone a significant evolution with the integration of causal inference. Initially, spatial analyses were predominantly descriptive—mapping poverty rates to visualize geographic patterns—or focused on enhancing predictive accuracy by incorporating spatial dependencies, as seen in efforts to model poverty outcomes without explicitly probing their causes. 

\subsection{The Rise of Spatial Causality}

As causal inference emerged as a central pillar of econometrics, it prompted a reorientation of spatial analysis toward understanding the causal mechanisms driving these spatial patterns. This shift gave rise to methodologies that leverage geographic structures to isolate causal effects. For example, spatial difference-in-differences \citep{dube2014spatial} adjusts for proximity effects to estimate policy impacts more precisely, while geographic instrumental variables (IV) and regression discontinuity designs exploit spatial boundaries—such as administrative borders or natural features—to identify causal relationships \citep{keele2015geographic, kelejian2004instrumental, betz2020spatial}. Furthermore, techniques like spatial difference-in-discontinuities \citep{butts2021difference} combine spatial and temporal dimensions to enhance causal identification. These works reflect a synthesis of spatial and causal thinking in spatial econometrics. 

In summary, the geography of poverty has witnessed an increase in quantitative approaches employing spatial estimation modeling. A significant limitation of these methods is their reliance on pre-specified parametric (functional) forms of distance-defined spatial dependencies, which struggle to capture unknown or complex non-linearities in the data. In the subsequent section, we examine recent trends incorporating more flexible modeling approaches for poverty prediction using machine learning that relax this assumption. 

\section{Earth Observation, Machine Learning, and Poverty Prediction}\label{s:ML}

Although traditional spatial statistical methods have advanced poverty analysis by modeling geographic dependencies, their reliance on a potentially rigid functional form involving distance-based relationships often fails to capture highly non-linear dynamics or longer-range dependencies in geospatial data. To address this, researchers have integrated EO data—such as satellite imagery—with ML and computer vision techniques, giving rise to EO-ML methods. ML is the scientific field studying how algorithms can learn from data automatically. These approaches extract detailed visual features, including road networks, building density, and agricultural land use, directly from imagery, enabling a more precise and nuanced mapping of economic conditions. In this way, addressing some of the constraints of distance-based models, EO-ML methods provide a framework for high-resolution poverty prediction.

Image-based poverty prediction is perhaps the most prominent use of current EO-ML methods in the geography of poverty. Whereas sub-national poverty maps discussed in Section \ref{s:ClassicalStat} have primarily been collected by governments at aggregated administrative levels of analysis (e.g., state or prefecture-level), EO-ML methods can produce maps with details at the pixel level. Although early studies used nighttime lights to estimate development without the ML component \citep{doll2010estimating}, more recent research combines ML with temporal and multi-spectral images \citep{pettersson2023time,babenko2017poverty,jean2016combining, daoud_using_2023-1, blumenstock2015predicting, yeh2020using}, and other spatial data \citep{marty2024global}. Because of the global availability of EO data, with the right training data, the EO-ML approach enables the imputation of areas without adequate survey data, where missingness can be present due to civil conflict \citep{gustafsson2024estimating} or natural disasters \citep{kakooei2022remote}. This strand of EO-ML research has been shown to explain 70\% \citep{yeh2020using} and 75\% \citep{pettersson2023time} of the variation in ground truth data. Also, outcomes vary in their predictiveness---``hard'' indicators such as roofing materials or roads are easier to detect compared to ``soft'' welfare indicators such as income or consumption \citep{hall2023review}.  While observing direct human interactions or behaviors from EO data is difficult with free and publicly available imagery, the physical changes that result can be measured---such as the construction of new houses, mining activities, or agricultural production---can be used as indirect measurements. 

In EO-ML poverty prediction, analysis takes the form of estimating the expected value $\mathbb{E(\cdot)}$ of poverty: 

\begin{align*} 
\mathbb{E}\left[ Y_i \, \mid \, \bM_i = \mathbf{m}\right] = f_{\boldsymbol{\theta}}(\bM) + \epsilon
\end{align*} 
\noindent where $Y_i$ denotes the poverty outcome in question, $f_{\boldsymbol{\theta}}$ denotes a machine learning model defined with parameters $\boldsymbol{\theta}$ (often in the millions) and $\bM_i$ denotes EO image arrays. In other words, we use the EO image data to form predictions of poverty for, e.g., households or in different grid cells. Whereas the spatial statistics literature models outcomes assuming interdependence between spatial units proportional to the distance between them, in EO-ML modeling, the associations between different spatial locations in the image arrays are allowed to flexibly interact with all other locations via computer vision techniques such as convolutions or attention mechanisms. In classical image processing for machine learning, convolutions use a small matrix, known as a kernel, to slide over an image, extracting features like edges or textures by combining pixel information across image neighborhoods. Additionally, the attention mechanism improves the model's ability to focus on key image regions by assigning greater weight to more relevant areas, improving performance in tasks such as object detection and image captioning.

Some of these EO-ML applications use transfer learning methods in $f_{\boldsymbol{\theta}}$ \citep{daoud_using_2023-1, xie2016transfer}. In such transfer learning methods, computer vision models using convolutional models or Vision Transformer backbones trained for other purposes are adapted to the prediction of poverty \citep{khan2022transformers}. For example, \citet{ni2020investigation} leverage transfer learning approaches, using pre-trained deep learning models (VGG-Net, Inception-Net, ResNet, and DenseNet) to extract features from daytime satellite imagery, which are then used as inputs in a lasso regression model to predict poverty levels.\footnote{A lasso regression model is equivalent to an OLS model that seeks to minimize the sum of squared residuals plus a penalty term for large coefficients.} In a related approach, \citet{rolf2021generalizable} introduced a method using randomized feature mappings to create vector representations of satellite imagery. This approach runs a randomly initialized single-layer convolutional network to obtain a large number of image features. They then use ridge regression to make predictions. This technique avoids the high computational cost of transfer learning models trained for other purposes while still yielding good performance.

To supplement these traditional ML approaches where labeled poverty data (e.g., from DHS surveys) are used to form poverty predictions with EO data, there is a rapidly growing set of EO-based foundation models for generating representations of image data that can be used in downstream poverty prediction tasks \citep{tsaris2024pretraining,ClayFoundationModel}. In contrast to the supervised approach, EO foundation model methods often use self-supervised techniques wherein patches of input satellite images are randomly masked (i.e., held out); the rest of the image is then fed through a vision model (e.g., Vision Transformers using a mechanism known as attention map that models the relationship between all parts of an image with all others) to predict the masked image portion \citep{badrinarayanan2017segnet}. A benefit of using EO foundation models for poverty analysis is that predictions of poverty based on image information from one location can be improved by leveraging large amounts of image data from outside the study area, as these foundation models are trained on terabytes of global image information.

In tandem with the rise of EO foundation models, researchers have begun to leverage the temporal aspect of satellite imagery. EO data streams contain temporal slices on a weekly (e.g., in Landsat images) to daily (e.g., in Maxar commercial images) basis, allowing for complex patterns of development and changes across the built environment to be modeled in generating a latent understanding of the characteristics of the place under consideration. By incorporating temporal dependence in models, a higher degree of accuracy can be achieved, as illustrated in \citep{pettersson2023time}. An implication of this work is that satellite time sequences reveal information about the progression of economic development, enabling improved predictions about the level of development at any given time slice. 

There are several important considerations when performing EO-ML analysis with spatial sensitivity. First, in a comparison of various ML methods against ordinary least squared regression, neural networks performed slightly better on prediction and reducing spatial autocorrelation of residuals \citep{song2023three}, that is, the systematic mis-prediction of outcomes in a spatially clustered way.\footnote{Convolutional methods were not included, but non-linear models seem to be more performant.} Still, research needs to be conducted to understand where residual spatial autocorrelation proliferates---and to probe whether this residual autocorrelation is due to data structure or model choice. To explore this question further, there are techniques like blocked cross-validation which can deal with different interactions among the units of interest (e.g., temporal, spatial, group, and hierarchical structures \citep{roberts2017cross}), yielding more realistic assessments of out-of-sample performance.

Beyond spatial residual autocorrelation in EO-ML model pipelines, the choice of EO image representation remains a crucial consideration that can significantly impact downstream analyses of neighborhood poverty prediction. For example, while foundation models are promising in that they are trained using a large corpus of satellite images (unlike random projection and many transfer learning approaches), they are not only expensive to train \citep{Smith2023NoFreeAILunch}, but also not specifically tuned to capture socioeconomic dynamics. Transfer learning approaches are limited by the fact that the original models are typically trained on mass media images rather than satellite imagery, which reduces their predictive power. In addition, randomized image representations are computationally efficient and do not require the fitting of complex models, but they are also limited in their ability to represent intricate associations between images and, e.g., spatial poverty outcomes. In this way, there are several trade-offs in the EO-ML prediction of poverty that researchers must grapple with in applied work.

In summary, while the literature on using EO-ML models to predict poverty is gaining traction, it remains unclear to what extent EO-ML models are useful for causal inference. The next section shows our findings with regard to that question. 

\section{A Scoping Review}\label{s:Scoping}

To evaluate the scope of EO-ML models for causal inference, we conducted a review of papers that met the following criteria: (1) utilized EO data, (2) employed a causal inference framework, (3) used ML methods, and (4) addressed a social science question. By causal inference, we refer to the statistical analysis of cause and effect, incorporating both design-based studies (e.g., randomization of the intervention) and observational studies \citep{imbens2015causal}.\footnote{This review excludes studies that use EO data solely for outcome measurement rather than for estimating causal effects.} Our search terms include poverty-related phrases, but also more general social science topics as well
in order to capture a wider range of methodological innovations and applications of EO-ML methods in causal inference. These innovations can also inform poverty research by highlighting transferable techniques that can also be applied to the geography of poverty. Moreover, the inclusion of other topics is motivated by how poverty intersects with a wide range of other socio-economic processes (e.g., environmental degradation, public health outcomes, resource dependence, displacement, and access to infrastructure). Our search phrases are outlined in the Appendix. 

\begin{figure}[H]
    \centering
\includegraphics[width=0.75\linewidth]{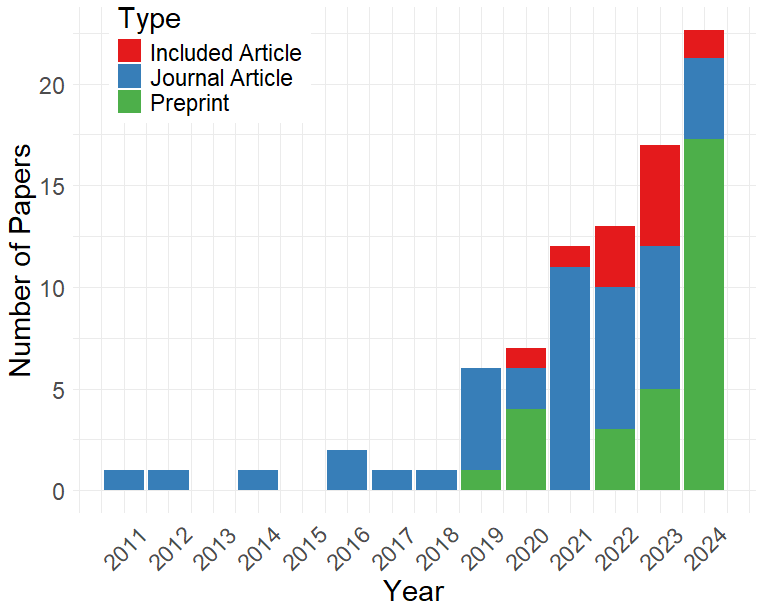}
\caption{Trends in research output on the intersection of causal inference, machine learning, and Earth observation.}
\label{fig:publicationYear}
\end{figure}

The information sources included Scopus, Web of Science, and IEEE Xplorer.\footnote{See \href{https://GitHub.com/AIandGlobalDevelopmentLab/eo-poverty-review}{\url{GitHub.com/AIandGlobalDevelopmentLab/eo-poverty-review}} for a list of papers found in the search.} The initial search yielded 101 papers published in journals and 41 pre-prints between 2011 and 2024 (up to August). We see in Figure ~\ref{fig:publicationYear} a summary of the publication and registration years for the sources. An average of 1 or fewer papers were published between 2011-2015, with an increase beginning in 2019. Although the total number of papers is still relatively scant compared to the overall volume of research in EO-ML methods for prediction, this rise indicates a growing interest in EO-ML methods for causal inference in social science contexts. 

\begin{figure}[H]
    \centering
\includegraphics[width=0.75\linewidth]{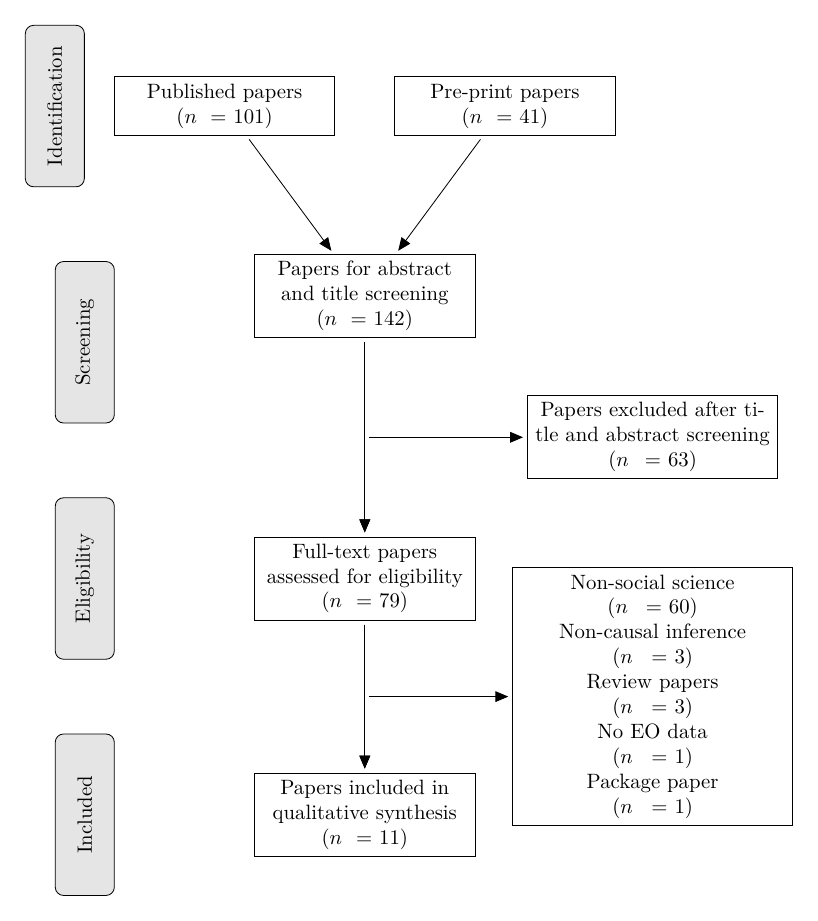}
\caption{Design for scoping literature review on causal EO-ML work.}
\label{fig:Eligibility}
\end{figure}

\begin{figure}[H]
\caption{
Summary of papers in the causal EO-ML literature. $*$ denotes a preprint (as of July 2024).
}\label{tab:PaperSummary}
  \centering
\includegraphics[width=0.99\linewidth]{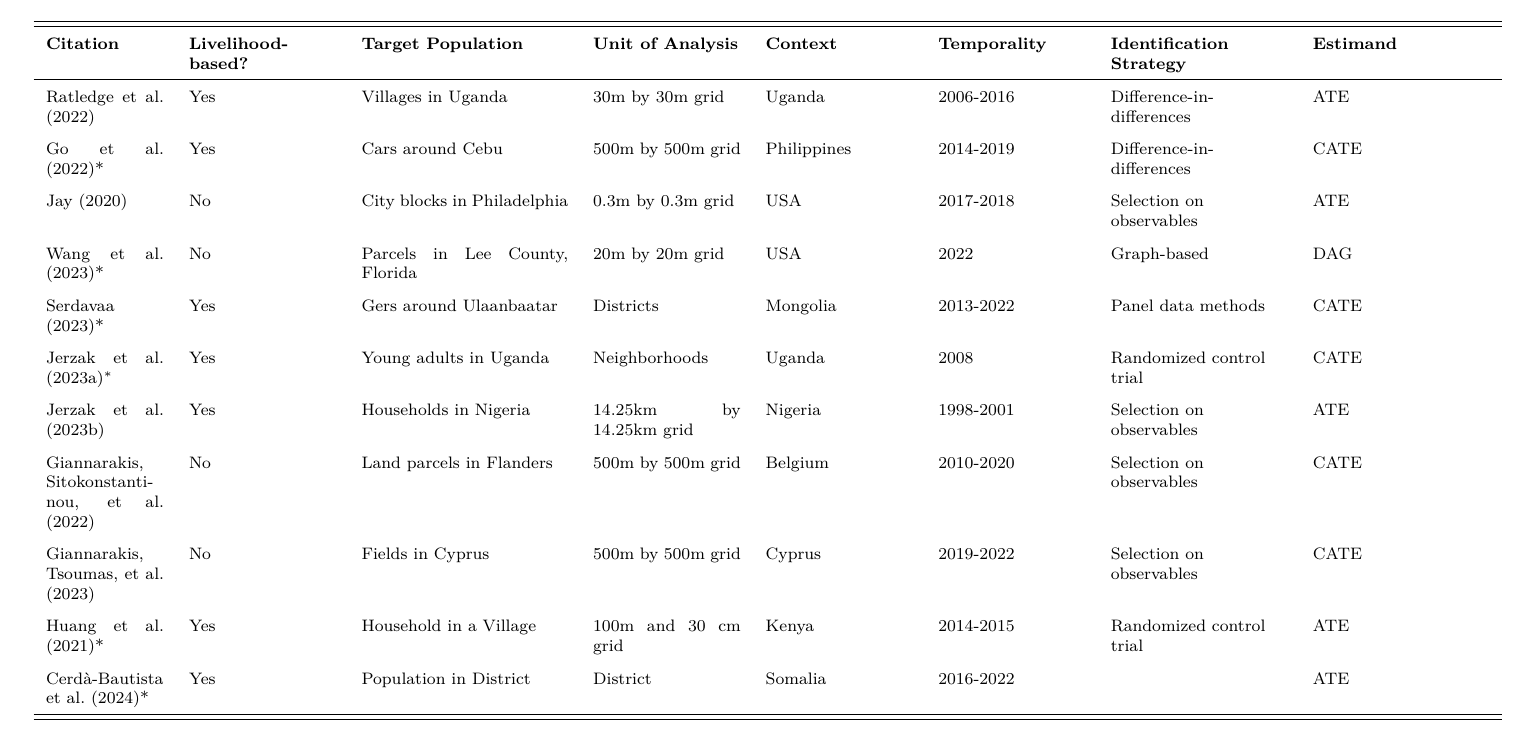}
\end{figure}

We constructed a schema to filter and synthesize the results from the query. The schema grouped sources by themes (domain, source type, aim, main findings, social science relevance, causal inference relevance), data attributes (outcome, exposure, temporality, context, unit of analysis, EO source(s), EO purpose(s), multiresolution, multiphase, and multisource), and methodology (causal estimand, identification strategy, causal estimator, and image model). These fields enable both the sorting of articles by causal inference approach and social science relevance. From the 142 queried sources, only 11 papers met the review requirements, shown in Table \ref{tab:PaperSummary}. Our initial screening of titles and abstracts excluded 63 papers, leaving 79 for full-text review. During the full-text review for eligibility based on the fields stated above, we excluded 68 papers based on: non-social science research ($n =$ 60), not being causal ($n =$ 3), being a review paper ($n =$ 3), not utilizing EO data ($n =$ 1), and a statistical package paper ($n =$ 1). We identified 11 papers for inclusion, 5 were published, 6 were pre-prints, and they were released between 2020 and 2023. The flowchart is presented in Figure \ref{fig:Eligibility}.

Thematically, the majority of papers assessed in full can be categorized into economics, environmental/Earth sciences, or methods development (causal discovery and estimation), with Earth/environmental sciences, being the most represented. We also find that compared with the 142 total papers collected, there are significantly fewer papers on social science topics, specifically poverty-related. This indicates that, as methodological foundations for machine learning-based causal inference with Earth observation data are being established, their integration into poverty research remains in its early stages.

The papers in our analysis set employed diverse methods for causal claims, as shown in Table~\ref{tab:PaperSummary}. Temporal approaches, like difference-in-differences (DiD) used by \citet{ratledge_using_2022} and \citet{go_use_2022}, exploit EO data’s time series to estimate average treatment effects (ATEs). \citet{serdavaa_satellite_2023} applied Granger causality, relying on stable variable relationships over time to suggest causation. Cross-sectional methods included \citet{jay_alcohol_2020}’s matching framework and \citet{jerzak2023integrating}’s Inverse Propensity Score Weighting (IPW), while \citet{cerda2024causal} combined IPW with T-learner and X-learner models for robustness. \citet{huang2021using} used simpler linear regression for ATEs. \citet{wang_causality-informed_2023} explored causal discovery to map variable relationships, and \citet{jerzak2023image} employed an experimental design. This range highlights EO-ML’s flexibility in advancing causal social science research.

On the geographic distribution of study areas, 5 focused on Africa, 2 on Asia, 2 on Europe, and the remaining 2 on the United States. While papers on Africa and Asia studied poverty or economic development issues, the papers on the United States and Europe did not. For example, \citep{jay_alcohol_2020} was concerned with a public health question on the causal effect of alcohol and firearm violence, while \citep{wang_causality-informed_2023} explored emergency management by studying building damage from Hurricane Ian in Florida. Focusing on Europe, \cite{giannarakis_towards_2022} and \citet{giannarakis_understanding_2023} were concerned with the net primary production based on various agricultural practices selected. 

Lastly, we find that the majority of papers in the causal space use EO data for outcome or covariate imputation, as opposed to deconfounding or heterogeneity decomposition purposes. The use of EO data for imputing outcomes and covariates seems to be the preferred method thus far in the causal EO-ML literature, possibly because it is the most straightforward way of incorporating EO data into the analysis. There is a protracted history of geographers using satellite images and raster-based analysis, but more recent ML methods, such as foundation models, are only beginning to be incorporated \citep{jerzak2024effect}. These methods and the adaption of image-based causal strategies have significant promise for EO data in the geography of poverty. 

We also observe that the papers face various degrees of inference problems discussed in \citet{meng2014trio}---problems related to multi-resolution, multi-phase, and multi-source considerations. Each challenge can reduce the signal-to-noise ratio in causal analysis. For example, if satellite imagery is at one resolution but official census or survey data is at another, potential bias is introduced.

\begin{figure}[H]
  \caption{Summary of papers in the causal EO-ML literature regarding multi-resolution, multi-phase, and multi-source considerations.
}\label{tab:MultiAnalysis}
  \centering
\includegraphics[width=0.99\linewidth]{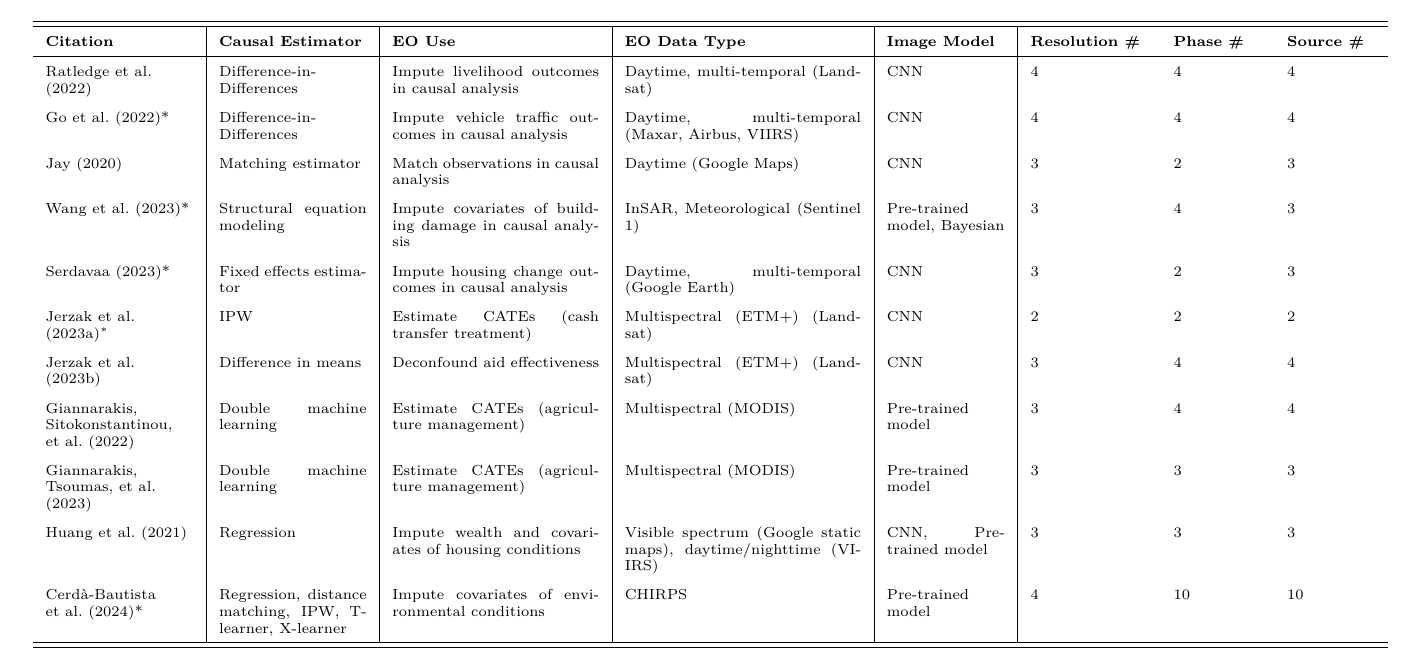}
\end{figure}

Moreover, the number of resolution mismatches can increase as additional sensor inputs are included; phase mismatches can occur as data products from one time period are used with data from another (e.g., decadal land use classification maps with raw satellite data). The number of sources can also grow as investigators include information from multiple organizations like the United Nations, World Bank, OECD, and USAID (which all produce distinct statistics with different sampling frames). We see how each paper manifests these inference problems in Table \ref{tab:MultiAnalysis}.

\section{A Protocol for EO-ML Causal Modeling}\label{s:protocol}

Based on our scoping review synthesis, we develop a protocol for applying EO-ML models for causal inference. This protocol serves as a general guideline for conducting causal EO-ML analysis, whether for outcome imputation, deconfounding, or experimental analysis. The primary audience for this chapter includes those interested in poverty research, though its methodologies can readily be applied to broader social science inquiries. We created Figure \ref{fig:Protocol}, which is a visual abstract summarized in eight key steps. These steps are, in turn, grouped into three broader stages: {\sc Question and Identification}, {\sc Earth Observation Data}, and {\sc Estimation and Uncertainty Quantification}. 

\subsection{Question and Identification}\label{s:CausalMLEO}

Our scoping review suggests five ways that EO-ML methods enable causal analysis: (1) poverty outcome imputation for downstream causal analysis, (2) EO image deconfounding, (3) EO-based treatment effect heterogeneity, (4) EO-based transportability analysis, and (5) satellite-derived causal discovery. While these five ways are not exhaustive of all possibilities, they likely cover a plurality of ways that social scientists would use EO-ML modeling for causal inference. In this section, we develop a general protocol for how to use EO-ML modeling for causal inference using these five approaches. 

The first approach in causal EO-ML methodology involves outcome imputation for downstream causal analysis. Indeed, scholars investigating the causes of poverty with EO data have, as an initial entry, used ideas from the previous section on satellite-based predictions of poverty with causal inference methods. Formally, when direct measurements of the outcome \( Y_i \) are unavailable, EO data can be used to impute \( Y_i \) via a machine learning model \( f_{\theta}(\mathbf{M}_i) \). This imputed outcome can then be used in causal estimation methods like difference-in-differences to estimate the ATE, with possible adjustment for imputation bias. 

For example, \citet{ratledge_using_2022} measured the impacts of electrification in Uganda by combining EO-ML predictions of local livelihood measures. They then used a difference-in-differences analysis to estimate the causal effect of electrification on livelihood. Another study combined EO-ML classification and enumeration of automobiles with difference-in-differences to measure change in traffic volumes following the construction of a new airport \citep{go_use_2022}. Here, the EO-derived outcomes create time series data that resonate with the difference-in-differences identification strategy---making it the most suitable strategy.  

The second approach utilizes EO-images for deconfounding. Because confounding is likely omnipresent in observational studies, a natural question to ask is ``Does EO data capture information about confounding?'' As satellite imagery contains information about a host of economic, geographic, and environmental factors, that confounding may be directly observed in an image (e.g., that aid projects are placed closer to cities) or latent (e.g., poor neighborhoods) \citep{jerzak2023integrating}. In both these cases, EO-ML models can be used to account for confounding by modeling the probability of the intervention, given the image: 
\[
g\left( {\Pr}(A_i \mid \bM_i = \bmm) \right) \ = \  \tilde{f}_{\boldsymbol{\theta}}(\bmm) + \epsilon, 
\]
\noindent where $g$ is any suitable link function (e.g., logit), $\tilde{f}_{\boldsymbol{\theta}}(\cdot)$ defines an ML model with parameters $\boldsymbol{\theta}$ relating the image or image sequence information, $\bM_i$ with the probability of treatment, $A_i$, and residual $\epsilon$. This probability can then be used in a variety of reweighting approaches, such as double machine learning or inverse probability weighting methods \citep{knaus2022double}.

The third approach harnesses EO data for treatment effect heterogeneity. We see that the causal EO-ML literature has begun using EO data in the analysis of treatment effects in experimental data. For example, \citet{jerzak2023image} estimates conditional average treatment effects (CATEs) using satellite data to model the kinds of neighborhoods most responsive to anti-poverty interventions. In context, there is substantive and policy interest in modeling 
\[
\tau(\bmm) = \mathbb{E}[Y_i(1) - Y_i(0) \mid \bM_i = \bmm],
\]
\noindent where $\bM_i$ represents a satellite image or image sequence representation around $i$'s location (where $i$ could be a household or a village), where $Y_i(1)$ represents the potential outcome under treatment, $Y_i(0)$ the potential outcome under control, and $Y_i(1) - Y_i(0)$ represents the difference in poverty outcomes with and without an economic intervention (such as a cash transfer). Substantively, $\tau(\bmm)$ can inform decision-makers about the kinds of neighborhood structures where individuals are most responsive to anti-poverty interventions. EO foundation models \citep{ClayFoundationModel}, randomized projections \citep{rolf2021generalizable}, and end-to-end ML architectures can also be used in translating the raw image array, $\bM_i$, into a useful representation for heterogeneity decomposition. 

The fourth approach extends causal findings beyond their original context through transportability analysis \citep{pearl2011transportability}. Consider a cash transfer program evaluated in one region: transportability allows us to predict its effects in regions where no experiment was conducted \citep{schwartz2011transportability}. This is possible because EO data provides consistent imagery across locations. Indeed, the model for $\tau(\bmm)$ discussed above can be applied, with or without re-weighting, to the area outside the original geographic context of the experiment \citep{degtiar2023review}: image data is available not only for experimental sites but also all others in a region of interest. Formally, \(\mathbf{M}_i\), acts as a globally accessible covariate capturing location-specific features that influence treatment effects. A machine learning model, \(h_{\theta}(\mathbf{M}_i)\), is trained on experimental data from the source population (for whom \(S_i=1\)) to estimate the heterogeneity process \citep{jerzak2024effect}:
\[
\hat{\tau}(\mathbf{M}_i) = h_{\theta}(\mathbf{M}_i)
\]
Under one set of transportability conditions:
\[
\mathbb{E}[Y_i(a) \mid \mathbf{M}_i, S_i=1] = \mathbb{E}[Y_i(a) \mid \mathbf{M}_i, S_i=0] \quad \text{for} \quad a = 0, 1
\]
The learned heterogeneity model can be applied to new locations using satellite imagery. Thus, EO data facilitates the generalization of causal effects across diverse geographic contexts by leveraging the invariance of the CATE model conditioned on satellite imagery. This kind of disaggregation can help in the discovery of individuals or groups harmed by an overall effective treatment for the development of policy optimization rules \citep{sachdeva2023geographical}. 

Here, some caution is warranted. This sort of transportability analysis hinges on the assumption that EO-derived features accurately reflect sources of variation in treatment effects. If the EO data fails to capture key moderators, the resulting subgroup analyses may misidentify which individuals or groups are adversely affected. Researchers should therefore exercise caution, ensuring that EO features are theoretically justified and empirically validated as proxies for the causal mechanisms at play.

The fifth and last approach is for \emph{causal discovery}, where the aim is to learn the underlying causal structure among multiple observed variables (including those derived from EO data). Formally, the goal is to identify a directed acyclic graph (DAG) $\mathcal{G}$ on a set of variables $\{X_{i1},\dots,X_{id}\}$ that best explains their joint distribution $p(X_{i1},\dots,X_{id})$, subject to assumptions such as acyclicity, the Markov property, and faithfulness \citep{pearl2009causality}.

Because satellite imagery can capture information about environmental, infrastructural, and other latent factors, it can be incorporated into the causal discovery pipeline to illuminate how these factors connect to socioeconomic outcomes \citep{runge2019inferring}. For example, \citet{ebert2012causal} uses discovery algorithms to propose relationships in satellite-based representations of spatial systems, thereby improving the credibility of causal insights in complex, observational settings.

In short, causal inference methods have been advanced to handle EO image arrays to model cause and effect related to neighborhood-level poverty. Some of this work has focused on using EO-ML imputed poverty outcomes in causal analyses \citep{ratledge_using_2022}; others have used EO data for treatment effect heterogeneity reasoning or for deconfounding \citep{jerzak2023image}. 

\begin{figure}[H]
  \centering
  \includegraphics[width=0.99\linewidth]{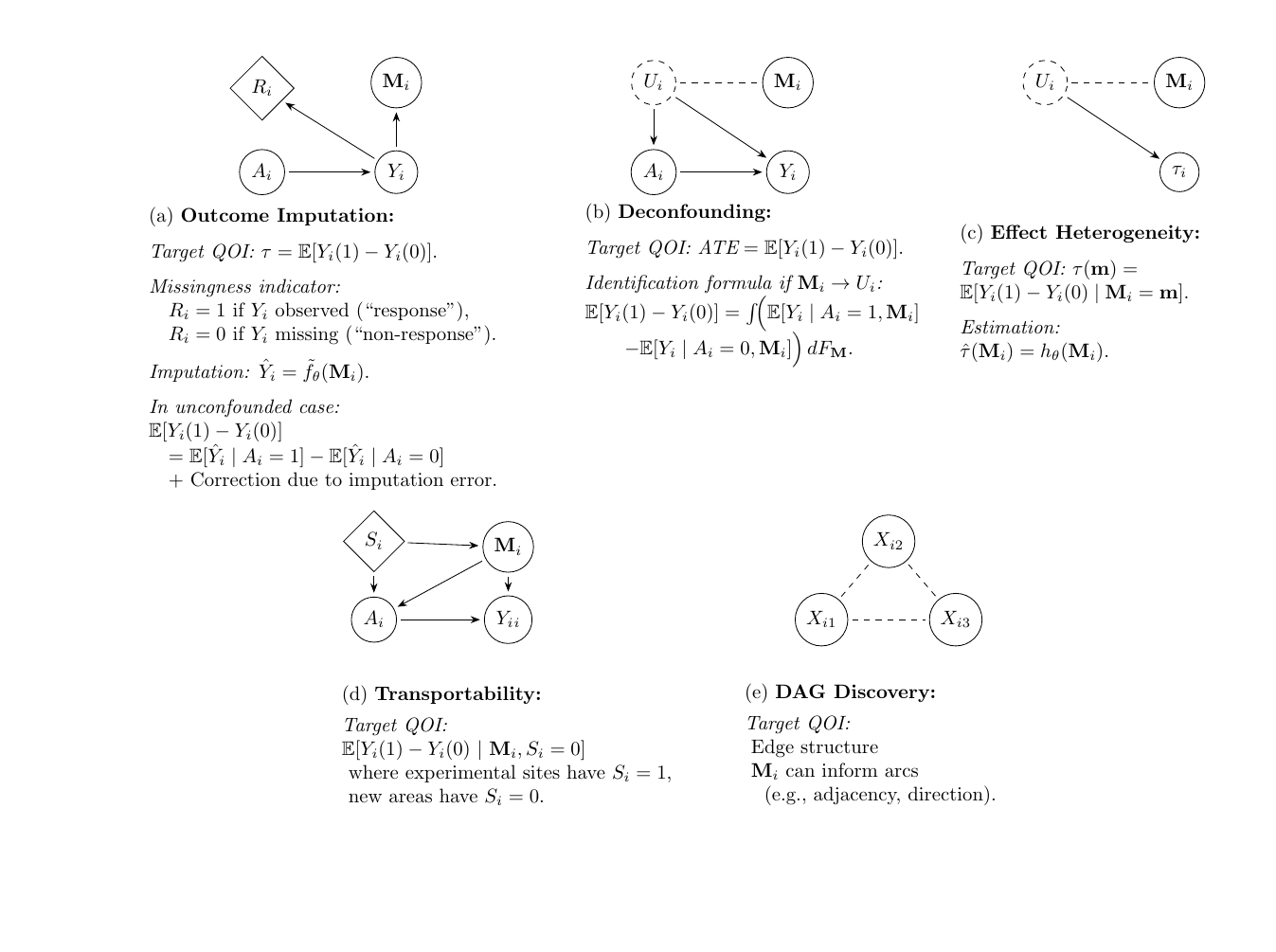}
  \caption{Base graphs of causal EO-ML methods. Images are often seen as a proxy of other important unobserved factors, $U_i$. $\tau_i$ denotes the causal effect of $A_i$ on $Y_i$. $\bM_i$ represents a satellite image array; $Y_i$ denotes the outcome, $A_i$ intervention of interest, and $U_i$ unobserved factors that help drive the causal system and that are indirectly captured in satellite image representations. Diamond nodes represent selection or missingness indicators. Dotted lines represent relationships of unknown or indeterminate directionality.
  }
\label{fig:ThreeDagsProxy}
\end{figure}

Figure \ref{fig:ThreeDagsProxy} illustrates these five applications. In these DAGs, those unobserved factors, \( U_i \), are represented by a dotted circle, and the dotted lines connecting \( U_i \) to other variables (e.g., \( \mathbf{M}_i \), \( A_i \), or \( Y_i \)) indicate indirect relationships where these unobserved factors affect the system but are not necessarily measurable directly. Instead, \( \mathbf{M}_i \) captures information about \( U_i \), such as urban/rural distinctions or neighborhood structures, which can be critical for addressing confounding in contexts like economic aid analysis. For example, satellite imagery might reveal patterns of infrastructure or land use that reflect underlying socio-economic conditions influencing both aid allocation and poverty outcomes.

The top left DAG (panel $a$) represents the \emph{outcome-imputation} scenario, where \( M_i \) is used to estimate or impute the outcome \( Y_i \) (e.g., poverty levels) from a small training set (e.g., of DHS data) to impute to a larger area (e.g., all of Africa). The middle DAG (panel $b$) depicts the \emph{deconfounding} setup, where \( M_i \) adjusts for confounding by providing insights into \( U_i \), which may affect both the intervention \( A_i \) (e.g., aid programs) and the outcome \( Y_i \). The top right DAG (panel $c$) visualizes the \emph{effect-heterogeneity} setup following the best practices for visualizing heterogeneity described in \cite{nilsson2021directed}; here, \( M_i \) helps explore how the causal effect of \( A_i \) on \( Y_i \) varies across different contexts, using \( U_i \) as a proxy, which also supports transportability analysis to extend findings to new settings.  The bottom rows correspond to the {\it transportability} and \textit{DAG discovery} situations, respectively.

With these five approaches in mind, we return to the broader framing of the {\it Question and Identification} part of the protocol, which contains four steps, which focus on the definition of the causal aim and clarifying how EO data support the achievement of that aim. Investigators will (1) define the causal research question by specifying the treatment of interest $A_i$---such as aid programs, economic interventions, and other policy interventions---and the outcome $Y_i$. In poverty research, that outcome can be peoples' material assets, consumption, or income. Investigators will (2) construct a causal diagram that outlines the potential interactions between variables, including the instrumental variables $Z_i$, confounders $X_i$, moderators, as well as the treatment $A_i$ and outcome $Y_i$ previously defined. Next, in (3), researchers will evaluate which of the defined variables could plausibly be measured in the EO data. This could include all variables, a partial selection of variables, or none. If none, the problem will not benefit from the EO data, but if there is at least one variable that can be measured, the focus should be on those variable(s) that are not measured better (with lower error) from tabular data. 

\begin{figure}[H]
    \centering
\includegraphics[width=0.9\linewidth]{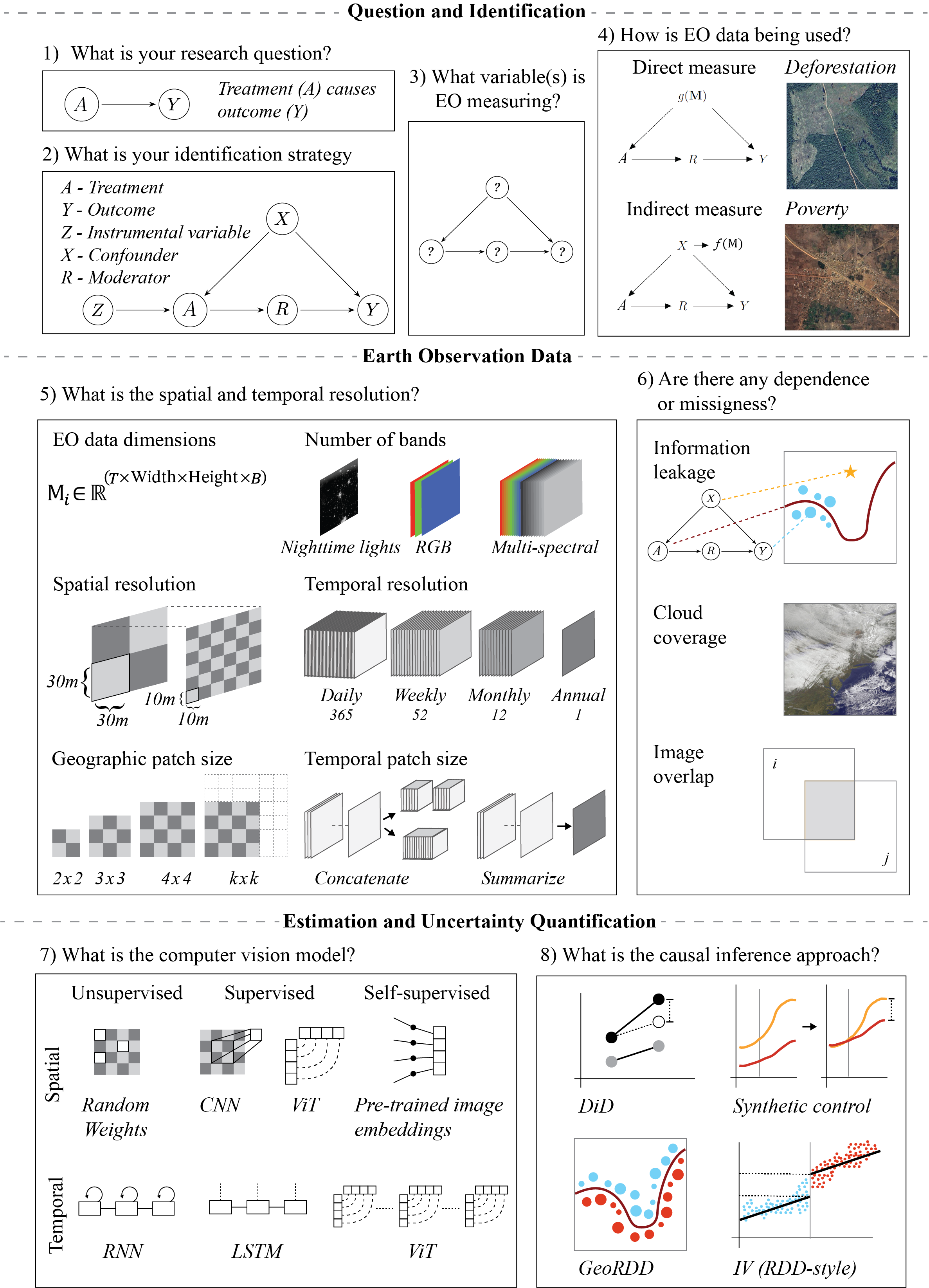}
    \caption{A visualization of a causal EO-ML protocol. ``CNN'' denotes Convolutional Network; ``LSTM'' denotes Long-Short-term Memory; ``RNN'' denotes Recurrent Neural Network; ``ViT'' denotes Vision Transformer; ``DiD'' denote difference-in-differences; ``Geo-RDD'' denotes geographic regression discontinuity.}
    \label{fig:Protocol}
\end{figure}

Finally, in (4), investigators will consider whether measurements from EO data will be directly quantifiable in the image and thus derived from pixel values (e.g., deforestation, land use, wildfire, etc.) or indirectly measured which require further processing (e.g., poverty, population, famine, NDVI, etc.). We define $g(\cdot)$ as a function that processes the EO data $\bM_i$ to produce a direct measure of the variable of interest, as observed in an EO image. This function can be an ML algorithm, a human annotator segmenting parts of an image, or both. In Figure \ref{fig:Protocol}, this $g(\bM_i)$ represents a measure of confounding (e.g., the level of deforestation in an area). In a direct measurement strategy, the assumption is that the measurement error is negligible; thus, we let $g(\bM_i)$ be a direct stand-in for the variable of interest. This is, of course, an assumption that only holds in ideal cases and thus should not be taken at face value. Conversely, we write an indirect measurement with an arrow to signify that measurement error is non-negligible. That is captured by $f(\bM_i)$, and exemplified by poverty being latently measured in EO images \citep{pettersson2023time}. 

\subsection{Earth Observation Data}\label{s:EOData}

The second component of the protocol guides the analyst in selecting and processing an appropriate EO dataset to measure the variable(s) defined in the previous step.  This protocol assumes that any other (tabular) data required for answering the research question is available. To measure the variable(s) of interest via an EO-ML approach, one must ask, ``What type of EO data and qualities do I need to measure the process or event of interest?''. 

In this vein, investigators must (5) define the appropriate resolution of the EO image $\bM_i$ and the EO bands, which jointly define the dimension of the EO data $\mathbb{R}^{T\times Width \times Height \times B}$. As previously mentioned, a band contains information about a specific wavelength of light. The index $B$ represents a numeric value specifying the number of frequency bands, $b = \{1,2,...,B\}$.  For example, to capture visible aspects of geography or human activity, the image must include the red, green, and blue (RGB) bands that make up a spectrum directly percevable by humans. Multispectral EO data contains these bands and also potentially others (e.g., short-wave bands measuring soil moisture), which recent research has used to measure poverty \citep{yeh2020using,pettersson2023time}. 

The unit of analysis can be defined by its temporal and spatial resolutions. \textit{Spatial resolution} refers to the level of detail captured by individual pixels in the image. This resolution is determined by the optical instrument mounted on the satellite. For example, Landsat satellites have an RGB resolution of 30-by-30 meters, while the same bands in Sentinel are captured at a 10-by-10-meter resolution, as illustrated in Figure \ref{fig:Protocol}. This difference corresponds to the size of the geographic area that any individual pixel can measure where a higher-resolution image from Sentinel enables a social scientist to see more details (as a Landsat pixel covers 900 $m^2$ [e.g., a typical suburban house lot], while a Sentinel pixel covers 100 $m^2$ [e.g., a large classroom in size], providing a resolution improvement of a factor of nine.)

\textit{Temporal resolution} corresponds to a satellite's revisiting frequency over an area. Public satellites have a lower revisit frequency (i.e., Landsat revisit the same area every two weeks; Sentinel satellites revisit every two to five days). Commercial satellites like Planet’s SkySat offer multiple daily revisits at 0.75 meters resolution.

If the spatial resolution defines the amount of detail in EO data, the \textit{geographic patch size} defines the dimensions of the image $\bM_i$. The $Width$ and $Height$ in $\mathbb{R}^{T\times Width \times Height \times B}$ specify the size of the patch, usually as a square. For example, an image with a patch size of 4-by-4 pixels and a resolution of 30 meters captures a square area of 120 meters on each side.  This image would cover a total geographical area of 14,400 $m^2$. Although remote sensing technology dictates image resolution, researchers can control patch size to fit the unit of analysis and surrounding area. 

Conversely, the \textit{temporal patch size} defines the timespan over which images are analyzed. Like spatial resolution, the temporal resolution is predetermined by the satellite technology selected, but the researcher can augment the size of the temporal patch, indexed by $t = \{1,...,T\}$, which is the final data dimension index of $\bM_i$. 

Concatenation and summarization provide distinct methods for modeling data at varying temporal resolutions and patch sizes. Concatenation takes two or more image arrays from adjacent periods and creates an image sequence array from these temporal slices. For example, if biweekly images are available, roughly 26 image snapshots per year are available in a given location, resulting in an array with 26 time slices ($T=$26).  

In contrast to direct concatenation, summarizing involves applying aggregating operations over pixels (or collections of pixels, which could alter the spatial resolution) to reduce data requirements and, consequently, computational burden. Such operations can include calculating the mean, median, variance, or other statistical quantities. There are other ways of model-based summarization involving, for example, wavelets that capture data changing image characteristics using signal processing methods \citep{kakooei2024analyzing}, convolutions, whether predefined or learned, or the most recent generation of end-to-end neural network methods. Selecting the appropriate way of summarizing temporal information requires knowledge about the rate of change one is analyzing. Like spatial resolution, there is a trade-off between how much temporal information can be reduced before the statistical signal deteriorates. 

In the last step (6), investigators will consider potential data issues, including information leakage, spatial dependence, image overlap, and missingness. Information leakage can introduce bias(es) in treatment effects since the EO image may contain statistical traces of several random variables specified in the causal diagram, but where the investigator only aims to use one (e.g., the confounder). This problem has been observed in text data \citep{tleakage2022}. For example, if the goal is to measure confounders in EO data (e.g., the existence of gold mines), but there are traces of the exposure (e.g., road construction programs) in the image, then using this data without removing these traces will create a biased estimate. The bias could increase even more if the image contains information traces about the outcome, as visualized in Figure \ref{fig:Protocol}. 

A known issue that arises in spatial analyses is \textit{spatial dependence} discussed in more detail in \ref{s:spatialeconometrics}  still applies to EO-ML modeling.  However, another challenge less discussed is image overlap. While spatial dependence is when two or more units have correlated outcomes, image overlap is when image patches overlap among two or more units. Such overlap will render the units dependent on their shared overlapping EO data. For example, if two units $i$ and $j$ have their respective images, we observe that the larger the overlapping area, the more likely the causal estimate could be biased. To reduce the risk of such image overlap, one could select smaller image patches. Yet, small patches may not represent the variable measured, thus, an exploratory data analysis is needed to identify the optimal trade-off between representation and image overlap. 

Lastly, while most EO sources have planetary coverage, collecting data from virtually every corner of the Earth, clouds can often create systematic missingness. For example, the cloudiest places (e.g., Torshavn in the Faroe Islands) have on average just 2.4 hours of sunlight per day, compared with Aswan, Egypt (having 10.6 hours) Although clouds hinder data collection in some key bands (e.g., RGB) which rely on passive sensors, many active sensors like synthetic-aperture radar (using microwave pulses) can penetrate clouds, but they typically provide lower spatial resolution and different types of information than optical sensors, making direct comparison or substitution challenging. Consequently, one standard way to handle missingness in EO data is to select the cloud-free image(s) or take these images and summarize the pixel value over a specified period. 

\subsection{Estimation and Uncertainty Quantification}\label{s:Estimation}

The last component is the {\it Estimation and Uncertainty Quantification}, where the methodological considerations will be decided upon. In step (7), the analyst evaluates how to process the EO data processed in the first phase. The data will have a spatial component but often a temporal dimension in addition, requiring the ML model to be specified. Computer vision offers a myriad of algorithms to process spatio-temporal EO data--from convolutional to attention-based methods \citep{vaswani2017attention}.

While \textit{supervised} ML trains on labeled outcomes like poverty, \textit{unsupervised} algorithms often focus on reducing the EO data, $\bM_i$, to lower-dimensional representations. For example, \citet{rolf2021generalizable} developed an unsupervised EO-ML method called Multi-task Observation using Satellite Imagery \& Kitchen Sinks. This method reduces storage and computational requirements by randomly sampling image patches and applying convolutions only to those patches, resulting in a compact representation of the original EO data. \textit{Self-supervised} methods such as those associated with the Clay Foundation model, further enhance these representations by pre-training on large EO datasets before downstream tasks using an auto encoder. Although computationally expensive to train, they are considered state-of-the-art. Ultimately, the goal is to find an effective computer vision representation of the variable defined in earlier steps \citep{prince2023understanding}.

With representations of $\bM_i$ in hand, we will proceed to the causal estimation stage (8) where one reflects on estimators and their properties. Depending on the research design and the identification strategy defined in the earlier stages, the researchers select the appropriate estimator. If the designs rely on temporal trends, then a DiD estimator or synthetic control are natural options. For example, \citet{ratledge_using_2022} used EO-ML methods to estimate the outcome of a neighborhood's living conditions by using DiD to evaluate the impact of electrification. If EO-ML representations are used for proxying confounding, like in \citet{jerzak2023integrating} or \cite{gordon2023remote}, then the estimator could be doubly robust or similar. 

While traditional statistical inference focuses on quantifying sampling variability through confidence intervals and standard errors, EO-ML models face additional challenges due to the high dimensionality of spatial data \citep{kakooei2024increasing}. Recent approaches addressing these challenges include prediction-powered inference \citep{angelopoulos2023prediction} and conformal prediction \citep{angelopoulos2021gentle}, though optimal uncertainty quantification for causal EO-ML models remains an active research area likely to see rapid developments soon.
  
\section{Discussion \& Future Directions}\label{s:FutureDirection}

Our scoping review has focused on causal EO-ML modeling for the geography of poverty. We discussed the links between traditional spatial statistical methods and EO-ML models. We argued that spatial statistical methods, while valuable in contributing to clear hypothesis-driven deductive research design, are limited in capturing non-linear dependencies in complex spatiotemporal EO data. This is what we discussed next: the rise of adapting ML methods to EO data, focusing on estimating poverty in satellite images \citep{jean2016combining,pettersson2023time,daoud_using_2023-1}. This area has seen rapid growth \citep{olofsson2014good} integrating other data sources, such as social media and mobile phone data, to improve poverty estimates \citep{blumenstock2015predicting,aiken2021machine}. EO-ML models have opened up new opportunities for researchers, complementing traditional approaches. Because EO-ML models have shown great promise in estimating poverty from images, the natural subsequent question is how well they work for causal questions. 

We next reviewed the current literature, both published and working papers where our search strategy identified 11 paper matches. Comparing this number to the size of the literature using spatial methods to analyze poverty, and the expanding articles with EO-ML methods for poverty prediction, we concluded that the use of causal EO-ML modeling is still nascent. This means that there are opportunities to make methodological and substantive contributions. The spatial coverage of EO data opens opportunities for new forms of comparative research over entire continents or the planet as a whole. We believe EO data can play an intermediary role to contribute to filling existing data gaps across countries and at fine-grained subnational scales while enabling new ways of analyzing global poverty.

Novel data sources and methodologies can expand our borders of knowledge but also face limitations. First, open questions remain about how to handle information leakage in EO-ML methods for causal inference. These are questions commonly addressed in causal representation learning \citep{scholkopf2021toward}, and we foresee research cross-fertilization across this field and EO-ML methods for social science research. Second, spatial dependence is prevalent \citep{akbari2023spatial}, and EO-ML methods need to address it. While some methods, such as the block bootstrap, offer a direct way of addressing stratified units and, thus, spatially dependent units, there is still room to develop best practices. Here, causal inference in social network analysis is another area for collaboration \citep{ogburn2024causal}.

Third, many open questions remain about EO foundation models and their role in causal inference \citep{dionelis2024evaluating}. Although foundation models may generate useful representations of EO data streams for causal heterogeneity or confounding analysis, weights from foundation models contain information from the entire training period of images (usually both pre- and post-treatment, especially for historical interventions). Using data that includes information from both pre-and post-treatment periods can lead to information leakage, particularly treatment leakage, which may bias causal estimates \citep{tleakage2022}. Moreover, how to optimally quantify uncertainty for causal EO-ML models is an active area of research, and new methods are emerging rapidly \cite{kakooei2024increasing,world2018mainstreaming}.

Fourth, EO-ML methods require greater interpretability to determine which elements of imagery are informative for causal studies. Earth observation data remains underutilized; while we have strong capabilities for analyzing land cover and land use, imagery contains a multitude of additional features waiting to be extracted. Some EO research has begun leveraging text for improved interpretability \citep{wang2024skyscript}. A promising research direction is adapting large language models for causal EO-ML. Such models could identify objects within images that moderate effects or serve as mechanisms driving observed outcomes. 

Fifth, building upon the theoretical development initiated by Sen's conceptualization of poverty \cite{sen1990development}, development and aid organizations have subsequently created comprehensive indices to reflect the multidimensional nature of poverty \citep{UNDP1990HDI,un2023MPI}. As conceptualizations of poverty have evolved, causal econometric and machine learning (EO-ML) methodologies should correspondingly update their poverty outcome measurements to incorporate multidimensional metrics. 

Lastly, future work should discuss privacy and fairness when remote sensing technologies are used for causal research. For example, personally identifiable or sensitive information, such as race or gender could be reconstructed from EO data that does not explicitly contain such information \citep{perkins2009satellite}. Thus, the EO data streams may have statistical associations with the forbidden variable requiring further anonymization or obscuration to protect individuals (e.g., \citep{ravfogel2020null}). These concerns could motivate the use of synthetic EO or population data streams to protect those, especially at the highest risk or margins \citep{yates2022evaluation}.


\printbibliography

@article{llorente2015social,
  title={{Social Media Fingerprints of Unemployment}},
  author={Llorente, Alejandro and Garcia-Herranz, Manuel and Cebrian, Manuel and Moro, Esteban},
  journal={PloS one},
  volume={10},
  number={5},
  pages={e0128692},
  year={2015},
  publisher={Public Library of Science San Francisco, CA USA}
}

@article{li2024mitigating,
  title={{Mitigating Social Biases of Pre-trained Language Models via Contrastive Self-Debiasing with Double Data Augmentation}},
  author={Li, Yingji and Du, Mengnan and Song, Rui and Wang, Xin and Sun, Mingchen and Wang, Ying},
  journal={Artificial Intelligence},
  pages={104143},
  year={2024},
  publisher={Elsevier}
}

@article{papadogeorgou2022causal,
  title={{Causal Inference with Spatio-temporal Data: Estimating the Effects of Airstrikes on Insurgent Violence in Iraq}},
  author={Papadogeorgou, Georgia and Imai, Kosuke and Lyall, Jason and Li, Fan},
  journal={Journal of the Royal Statistical Society Series B: Statistical Methodology},
  volume={84},
  number={5},
  pages={1969--1999},
  year={2022},
  publisher={Oxford University Press}
}

@book{bedi2007more,
  title={{More Than a Pretty Picture: Using Poverty Maps to Design Better Policies and Interventions}},
  author={Bedi, Tara and Coudouel, Aline and Simler, Kenneth},
  year={2007},
  publisher={World Bank Publications}
}

@inproceedings{pettersson2023time,
  title={{Time Series of Satellite Imagery Improve Deep Learning Estimates of Neighborhood-level Poverty in Africa}},
  author={Pettersson, Markus B. and Kakooei, Mohammad and Ortheden, Julia and Johansson, Fredrik D. and Daoud, Adel},
  booktitle={Proceedings of the Thirty-Second International Joint Conference on Artificial Intelligence},
  pages={6165--6173},
  year={2023}
}

@article{marty2024global,
  title={{Global Poverty Estimation using Private and Public Sector Big Data Sources}},
  author={Marty, Robert and Duhaut, Alice},
  journal={Scientific Reports},
  volume={14},
  number={1},
  pages={3160},
  year={2024},
  publisher={Nature Publishing Group UK London}
}

@article{doll2010estimating,
  title={{Estimating Rural Populations Without Access to Electricity in Developing Countries through Night-time Light Satellite Imagery}},
  author={Doll, Christopher NH and Pachauri, Shonali},
  journal={Energy policy},
  volume={38},
  number={10},
  pages={5661--5670},
  year={2010},
  publisher={Elsevier}
}

@article{jean2016combining,
  title={{Combining Satellite Imagery and Machine Learning to Predict Poverty}},
  author={Jean, Neal and Burke, Marshall and Xie, Michael and Davis, W Matthew and Lobell, David B and Ermon, Stefano},
  journal={Science},
  volume={353},
  number={6301},
  pages={790--794},
  year={2016},
  publisher={American Association for the Advancement of Science}
}

@misc{ClayFoundationModel,
  author = {{Clay Foundation}},
  title = {{GitHub Model Repository}},
  year = {2023},
  howpublished = {\url{https://github.com/Clay-foundation/model}},
  note = {Accessed: 2023-05-01}
}

@article{badrinarayanan2017segnet,
  title={{Segnet: A Deep Convolutional Encoder-decoder Architecture for Image Segmentation}},
  author={Badrinarayanan, Vijay and Kendall, Alex and Cipolla, Roberto},
  journal={IEEE Transactions on Pattern Analysis and Machine Intelligence},
  volume={39},
  number={12},
  pages={2481--2495},
  year={2017},
  publisher={IEEE}
}

@article{tsaris2024pretraining,
  title={{Pretraining Billion-scale Geospatial Foundational Models on Frontier}},
  author={Tsaris, Aristeidis and Dias, Philipe Ambrozio and Potnis, Abhishek and Yin, Junqi and Wang, Feiyi and Lunga, Dalton},
  journal={arXiv preprint arXiv:2404.11706},
  year={2024}
}

@misc{Smith2023NoFreeAILunch,
  author = {Smith, Craig},
  title = {{What Large Models Cost You -- There Is No Free AI Lunch}},
  year = {2023},
  note = {Accessed: 2024-05-02}
}

@article{ni2020investigation,
  title={{An Investigation on Deep Learning Approaches to Cmbining Nighttime and Daytime Satellite Imagery for Poverty Prediction}},
  author={Ni, Ye and Li, Xutao and Ye, Yunming and Li, Yan and Li, Chunshan and Chu, Dianhui},
  journal={IEEE Geoscience and Remote Sensing Letters},
  volume={18},
  number={9},
  pages={1545--1549},
  year={2020},
  publisher={IEEE}
}

@book{jalan1997spatial,
  title={{Spatial Poverty Traps?}},
  author={Jalan, Jyotsna and Ravallion, Martin and others},
  number={1862},
  year={1997},
  publisher={World Bank, Development Research Group Washington, DC, USA}
}

@article{blumenstock2015,
author = {Joshua E. Blumenstock},
title = {{Fighting Poverty with Data}},
journal = {Science},
volume = {353},
number = {6301},
pages = {753-754},
year = {2016},
doi = {10.1126/science.aah5217},
URL = {https://www.science.org/doi/abs/10.1126/science.aah5217},
eprint = {https://www.science.org/doi/pdf/10.1126/science.aah5217}}

@article{sen1990development,
  title={{Development as Capability Expansion}},
  author={Sen, Amartya},
  journal={The community development reader},
  volume={41},
  pages={58},
  year={1990}
}

@article{yang2023ecological,
  title={{Ecological and Social Poverty Traps: Complex Interactions Moving Toward Sustainable Development}},
  author={Yang, Zhenshan and Yang, Ding and Sun, Dongqi and Zhong, Linsheng},
  journal={Sustainable Development},
  volume={31},
  number={2},
  pages={853--864},
  year={2023},
  publisher={Wiley Online Library}
}

@article{burke2008spatial,
  title={{Spatial Disadvantages or Spatial Poverty Traps: Household Evidence from Rural Kenya}},
  journal={RePEc},
  author={Burke, William J. and Jayne, Thomas S.},
  year={2008}
}

@article{powell2001towards,
  title={{Towards a Geography of People Poverty and Place Poverty}},
  author={Powell, Martin and Boyne, George and Ashworth, Rachel},
  journal={Policy \& Politics},
  volume={29},
  number={3},
  pages={243--258},
  year={2001},
  publisher={Policy Press}
}

@inproceedings{tleakage2022,
    title = {{Conceptualizing Treatment Leakage in Text-based Causal Inference}},
    author = "Daoud, Adel  and
      Jerzak, Connor T.  and
      Johansson, Richard",
    editor = "Carpuat, Marine  and
      de Marneffe, Marie-Catherine  and
      Meza Ruiz, Ivan Vladimir",
    booktitle = "Proceedings of the 2022 Conference of the North American Chapter of the Association for Computational Linguistics: Human Language Technologies",
    month = jul,
    year = "2022",
    address = "Seattle, United States",
    publisher = "Association for Computational Linguistics",
    url = "https://aclanthology.org/2022.naacl-main.413",
    doi = "10.18653/v1/2022.naacl-main.413",
    pages = "5638--5645"
}

@article{glaeser2011cities,
  title={{Cities, Productivity, and Quality of Life}},
  author={Glaeser, Edward},
  journal={Science},
  volume={333},
  number={6042},
  pages={592--594},
  year={2011},
  publisher={American Association for the Advancement of Science}
}

@article{chetty2014land,
  title={{Where is the Land of Opportunity? The Geography of Intergenerational Mobility in the United States}},
  author={Chetty, Raj and Hendren, Nathaniel and Kline, Patrick and Saez, Emmanuel},
  journal={The Quarterly Journal of Economics},
  volume={129},
  number={4},
  pages={1553--1623},
  year={2014},
  publisher={MIT Press}
}

@book{lobao2007sociology,
  title={{The Sociology of Spatial Inequality}},
  author={Lobao, Linda M and Hooks, Gregory and Tickamyer, Ann R},
  year={2007},
  publisher={Suny Press}
}

@techreport{bird2010spatial,
title={{Spatial Poverty Traps: An Overview}},
author={Bird, Kate and Higgins, Kate and Harris, Dan},
year={2010},
institution={Overseas Development Institute},
type={ODI Working Paper},
number={321},
url={http://cdn-odi-production.s3.amazonaws.com/media/documents/5514.pdf},
note={CPRC Working Paper 161}
}

@article{blumenstock2015predicting,
  title={{Predicting Poverty and Wealth from Mobile Phone Metadata}},
  author={Blumenstock, Joshua E. and Cadamuro, Gabriel and On, Robert},
  journal={Science},
  volume={350},
  number={6264},
  pages={1073--1076},
  year={2015},
  publisher={American Association for the Advancement of Science}
}

@article{bird2019addressing,
  title={{Addressing Spatial Poverty Traps}},
  author={Bird, Kate},
  journal={Chronic Poverty Advisory Network, London},
  volume={2},
  year={2019}
}

@article{jerzak2023image,
  title={Image-based Treatment Effect Heterogeneity},
  author={Jerzak, Connor T. and Fredrik D. Johansson and Adel Daoud},
  journal={Proceedings of the Second Conference on Causal Learning and Reasoning (CLeaR), Proceedings of Machine Learning Research (PMLR)},
  year={2023},
  volume={213},
  pages={531-552},
  publisher={}
}

@techreport{huang2021using,
  title={{Using Satellite Imagery and Deep Learning to Evaluate the Impact of Anti-poverty Programs}},
  author={Huang, Luna Yue and Hsiang, Solomon M and Gonzalez-Navarro, Marco},
  year={2021},
  institution={National Bureau of Economic Research}
}

@misc{serdavaa_satellite_2023,
	address = {Rochester, NY},
	type = {{SSRN} {Scholarly} {Paper}},
	title = {A {Satellite} {Image} {Analysis} on {Housing} {Conditions} and the {Effectiveness} of the {Affordable} {Housing} {Mortgage} {Program} in {Mongolia}: {A} {Deep} {Learning} {Approach}},
	shorttitle = {A {Satellite} {Image} {Analysis} on {Housing} {Conditions} and the {Effectiveness} of the {Affordable} {Housing} {Mortgage} {Program} in {Mongolia}},
	url = {https://papers.ssrn.com/abstract=4664966},
	doi = {10.2139/ssrn.4664966},
	language = {en},
	urldate = {2024-04-16},
	author = {Serdavaa, Batkhurel},
	month = dec,
	year = {2023},
}

@misc{wang_causality-informed_2023,
	title = {Causality-informed {Rapid} {Post}-hurricane {Building} {Damage} {Detection} in {Large} {Scale} from {InSAR} {Imagery}},
	url = {http://arxiv.org/abs/2310.01565},
	urldate = {2024-04-16},
	publisher = {arXiv},
	author = {Wang, Chenguang and Liu, Yepeng and Zhang, Xiaojian and Li, Xuechun and Paramygin, Vladimir and Subgranon, Arthriya and Sheng, Peter and Zhao, Xilei and Xu, Susu},
	month = oct,
	year = {2023},
	note = {arXiv:2310.01565 [cs, eess]
version: 1},
}

@misc{jerzak_causalimages_2023,
	title = {{CausalImages}: {An} {R} {Package} for {Causal} {Inference} with {Earth} {Observation}, {Bio}-medical, and {Social} {Science} {Images}},
	shorttitle = {{CausalImages}},
	url = {http://arxiv.org/abs/2310.00233},
	urldate = {2024-04-16},
	publisher = {arXiv},
	author = {Jerzak, Connor T. and Daoud, Adel},
	month = nov,
	year = {2023},
	note = {arXiv:2310.00233 [cs, stat]
version: 3},
}

@misc{giannarakis_understanding_2023,
	title = {{Understanding the Impacts of Crop Diversification in the Context of Climate Change: A Machine Learning Approach}},
	shorttitle = {Understanding the impacts of crop diversification in the context of climate change},
	url = {http://arxiv.org/abs/2307.08617},
	urldate = {2024-04-16},
	publisher = {arXiv},
	author = {Giannarakis, Georgios and Ilias Tsoumas and Neophytides, Stelios and Papoutsa, Christiana and Kontoes, Charalampos and Hadjimitsis, Diofantos},
	month = jul,
	year = {2023},
	note = {arXiv:2307.08617 [cs, q-bio]
version: 1},
}

@misc{go_use_2022,
	address = {Rochester, NY},
	type = {{SSRN} {Scholarly} {Paper}},
	title = {On the {Use} of {Satellite}-{Based} {Vehicle} {Flows} {Data} to {Assess} {Local} {Economic} {Activity}: {The} {Case} of {Philippine} {Cities}},
	shorttitle = {On the {Use} of {Satellite}-{Based} {Vehicle} {Flows} {Data} to {Assess} {Local} {Economic} {Activity}},
	url = {https://papers.ssrn.com/abstract=4057690},
	doi = {10.2139/ssrn.4057690},
	language = {en},
	urldate = {2024-04-16},
	author = {Go, Eugenia and Nakajima, Kentaro and Sawada, Yasuyuki and Taniguchi, Kiyoshi},
	month = mar,
	year = {2022},
}

@article{perez-suay_causal_2019,
	title = {Causal {Inference} in {Geoscience} and {Remote} {Sensing} {From} {Observational} {Data}},
	volume = {57},
	issn = {1558-0644},
	url = {https://ieeexplore.ieee.org/document/8475013},
	doi = {10.1109/TGRS.2018.2867002},
	number = {3},
	urldate = {2024-04-16},
	journal = {IEEE Transactions on Geoscience and Remote Sensing},
	author = {Pérez-Suay, Adrián and Camps-Valls, Gustau},
	month = mar,
	year = {2019},
	note = {Conference Name: IEEE Transactions on Geoscience and Remote Sensing},
	pages = {1502--1513},
}

@inproceedings{giannarakis_towards_2022,
	title = {Towards assessing agricultural land suitability with causal machine learning},
	url = {https://ieeexplore.ieee.org/document/9856946},
	doi = {10.1109/CVPRW56347.2022.00150},
	urldate = {2024-04-16},
	booktitle = {2022 {IEEE}/{CVF} {Conference} on {Computer} {Vision} and {Pattern} {Recognition} {Workshops} ({CVPRW})},
	author = {Giannarakis, Georgios and Sitokonstantinou, Vasileios and Lorilla, Roxanne Suzette and Kontoes, Charalampos},
	month = jun,
	year = {2022},
	note = {ISSN: 2160-7516},
	pages = {1441--1451},
}

@article{jay_alcohol_2020,
	title = {Alcohol outlets and firearm violence: a place-based case-control study using satellite imagery and machine learning},
	volume = {26},
	issn = {1353-8047, 1475-5785},
	shorttitle = {Alcohol outlets and firearm violence},
	doi = {10.1136/injuryprev-2019-043248},
	language = {English},
	number = {1},
	urldate = {2024-04-16},
	journal = {INJURY PREVENTION},
	author = {Jay, Jonathan},
	month = feb,
	year = {2020},
	note = {Num Pages: 6
Place: London
Publisher: BMJ Publishing Group
Web of Science ID: WOS:000514665600011},
	pages = {61--66},
}

@article{ratledge_using_2022,
	title = {{Using Machine Learning to Assess the Livelihood Impact of Electricity Access}},
	volume = {611},
	issn = {0028-0836, 1476-4687},
	url = {},
	doi = {10.1038/s41586-022-05322-8},
	language = {English},
	number = {7936},
	urldate = {2024-04-16},
	journal = {Nature},
	author = {Ratledge, Nathan and Cadamuro, Gabe and de la Cuesta, Brandon and Stigler, Matthieu and Burke, Marshall},
	month = nov,
	year = {2022},
	note = {Num Pages: 23
Place: Berlin
Publisher: Nature Portfolio
Web of Science ID: WOS:000884842400024},
	pages = {491--+},
}

@article{runge_causal_2023,
	title = {{Causal Inference for Time Series}},
	volume = {4},
	issn = {2662-138X},
	url = {},
	doi = {10.1038/s43017-023-00431-y},
	language = {English},
	number = {7},
	urldate = {2024-04-16},
	journal = {Nature Reviews Earth \& Environment},
	author = {Runge, Jakob and Gerhardus, Andreas and Varando, Gherardo and Eyring, Veronika and Camps-Valls, Gustau},
	month = jul,
	year = {2023},
	note = {Num Pages: 19
Place: London
Publisher: Springernature
Web of Science ID: WOS:001017712700002},
	pages = {487--505},
}

@article{milbourne2010geographies,
  title={{The Geographies of Poverty and Welfare}},
  author={Milbourne, Paul},
  journal={Geography Compass},
  volume={4},
  number={2},
  pages={158--171},
  year={2010},
  publisher={Wiley Online Library}
}

@article{akbari2023spatial,
  title={{Spatial Causality: A Systematic Review on Spatial Causal Inference}},
  author={Akbari, Kamal and Winter, Stephan and Tomko, Martin},
  journal={Geographical Analysis},
  volume={55},
  number={1},
  pages={56--89},
  year={2023},
  publisher={Wiley Online Library}
}

@article{jerzak2023integrating,
  title={I{ntegrating Earth Observation Data into Causal Inference: Challenges and Opportunities}},
  author={Jerzak, Connor T. and Johansson, Fredrik D. and Daoud, Adel},
  journal={arXiv preprint arXiv:2301.12985},
  year={2023}
}

@article{ebert2012causal,
  title={{Causal Discovery for Climate Research Using Graphical Models}},
  author={Ebert-Uphoff, Imme and Deng, Yi},
  journal={Journal of Climate},
  volume={25},
  number={17},
  pages={5648--5665},
  year={2012},
  publisher={American Meteorological Society}
}

@article{yates2022evaluation,
  title={{Evaluation of Synthetic Aerial Imagery using Unconditional Generative Adversarial Networks}},
  author={Yates, Matthew and Hart, Glen and Houghton, Robert and Torres, Mercedes Torres and Pound, Michael},
  journal={ISPRS Journal of Photogrammetry and Remote Sensing},
  volume={190},
  pages={231--251},
  year={2022},
  publisher={Elsevier}
}

@article{ravfogel2020null,
  title={{Null It Out: Guarding Protected Attributes by Iterative Nullspace Projection}},
  author={Ravfogel, Shauli and Elazar, Yanai and Gonen, Hila and Twiton, Michael and Goldberg, Yoav},
  journal={arXiv preprint arXiv:2004.07667},
  year={2020}
}

@report{un2023MPI,
  title={{2023 Global Multidimensional Poverty Index (MPI): Unstacking Global Poverty: Data for High Impact Action}},
  author={UNDP},
  year={2023},
  publisher={UNDP (United Nations Development Programme}
}

@article{sachs2001geography,
  title={{The Geography of Poverty and Wealth}},
  author={Sachs, Jeffrey D and Mellinger, Andrew D and Gallup, John L},
  journal={Scientific American},
  volume={284},
  number={3},
  pages={70--75},
  year={2001},
  publisher={JSTOR}
}

@article{acemoglu2010africa,
  title={{Why is Africa Poor?}},
  author={Acemoglu, Daron and Robinson, James A.},
  journal={Economic History of Developing Regions},
  volume={25},
  number={1},
  pages={21--50},
  year={2010},
  publisher={Taylor \& Francis}
}

@article{getis2008history,
  title={{A History of the Concept of Spatial Autocorrelation: A Geographer's Perspective}},
  author={Getis, Arthur},
  journal={Geographical Analysis},
  volume={40},
  number={3},
  pages={297--309},
  year={2008},
  publisher={Wiley Online Library}
}

@article{wilson2014geographic,
  title={{Geographic Information Science \& Technology: Body of Knowledge 2.0 Project}},
  author={Wilson, John P},
  journal={Final Report: University Consortium for Geographic Information Science},
  year={2014}
}

@article{sachdeva2023geographical,
  title={{A Geographical Perspective on Simpson's Paradox}},
  author={Sachdeva, Mehak and Fotheringham, A Stewart},
  journal={Journal of Spatial Information Science},
  number={26},
  pages={1--25},
  year={2023}
}

@incollection{griffith2020some,
  title={{Some Guidelines for Specifying the Geographic Weights Matrix Contained in Spatial Statistical Models}},
  author={Griffith, Daniel A},
  booktitle={Practical handbook of spatial statistics},
  pages={65--82},
  year={2020},
  publisher={CRC press}
}

@article{degtiar2023review,
  title={{A Review of Generalizability and Transportability}},
  author={Degtiar, Irina and Rose, Sherri},
  journal={Annual Review of Statistics and Its Application},
  volume={10},
  pages={501--524},
  year={2023},
  publisher={Annual Reviews}
}

@article{hipp2016general,
  title={{General Theory of Spatial Crime Patterns}},
  author={Hipp, John R},
  journal={Criminology},
  volume={54},
  number={4},
  pages={653--679},
  year={2016},
  publisher={Wiley Online Library}
}

@article{briggs2018poor,
  title={{Poor Targeting: A Gridded Spatial Analysis of the Degree to which Aid Reaches the Poor in Africa}},
  author={Briggs, Ryan C},
  journal={World Development},
  volume={103},
  pages={133--148},
  year={2018},
  publisher={Elsevier}
}

@book{petrucci2003application,
  title={{The Application of a Spatial Regression Model to the Analysis and Mapping of Poverty}},
  author={Petrucci, Alessandra and Salvati, Nicola and Seghieri, Chiara},
  number={7},
  year={2003},
  publisher={Food \& Agriculture Org.}
}

@article{knaus2022double,
  title={{Double Machine Learning-based programme Evaluation Under Unconfoundedness}},
  author={Knaus, Michael C},
  journal={The Econometrics Journal},
  volume={25},
  number={3},
  pages={602--627},
  year={2022},
  publisher={Oxford University Press}
}

@report{UNDP1990HDI,
    author ={{UNDP}},
    title = {Human Development Report 1990: Concept and Measurement of Human Development},
    publisher = {UNDP (United Nations Development Programme)},
    year = {1990}
}

@article{kino2021scoping,
  title={{A Scoping Review on the Use of Machine Learning in Research on Social Determinants of Health: Trends and Research Prospects}},
  author={Kino, Shiho and Hsu, Yu-Tien and Shiba, Koichiro and Chien, Yung-Shin and Mita, Carol and Kawachi, Ichiro and Daoud, Adel},
  journal={SSM-population Health},
  volume={15},
  pages={100836},
  year={2021},
  publisher={Elsevier}
}

@article{hall2023review,
  title={A review of machine learning and satellite imagery for poverty prediction: Implications for development research and applications},
  author={Hall, Ola and Dompae, Francis and Wahab, Ibrahim and Dzanku, Fred Mawunyo},
  journal={Journal of International Development},
  volume={35},
  number={7},
  pages={1753--1768},
  year={2023},
  publisher={Wiley Online Library}
}

@article{karniadakis2021physics,
  title={{Physics-informed Machine Learning}},
  author={Karniadakis, George Em and Kevrekidis, Ioannis G and Lu, Lu and Perdikaris, Paris and Wang, Sifan and Yang, Liu},
  journal={Nature Reviews Physics},
  volume={3},
  number={6},
  pages={422--440},
  year={2021},
  publisher={Nature Publishing Group}
}

@article{athey2018impact,
  title={The impact of machine learning on economics},
  author={Athey, Susan and others},
  journal={The economics of artificial intelligence: An agenda},
  pages={507--547},
  year={2018}
}

@article{song2023three,
  title={Three Common Machine Learning Algorithms Neither Enhance Prediction Accuracy Nor Reduce Spatial Autocorrelation in Residuals: An Analysis of Twenty-five Socioeconomic Data Sets},
  author={Song, Insang and Kim, Daehyun},
  journal={Geographical Analysis},
  volume={55},
  number={4},
  pages={585--620},
  year={2023},
  publisher={Wiley Online Library}
}

@article{roberts2017cross,
  title={Cross-validation strategies for data with temporal, spatial, hierarchical, or phylogenetic structure},
  author={Roberts, David R and Bahn, Volker and Ciuti, Simone and Boyce, Mark S and Elith, Jane and Guillera-Arroita, Gurutzeta and Hauenstein, Severin and Lahoz-Monfort, Jos{\'e} J and Schr{\"o}der, Boris and Thuiller, Wilfried and others},
  journal={Ecography},
  volume={40},
  number={8},
  pages={913--929},
  year={2017},
  publisher={Wiley Online Library}
}

@article{fay1979estimates,
  title={Estimates of income for small places: an application of James-Stein procedures to census data},
  author={Fay III, Robert E and Herriot, Roger A},
  journal={Journal of the American Statistical Association},
  volume={74},
  number={366a},
  pages={269--277},
  year={1979},
  publisher={Taylor \& Francis}
}

@article{elbers2003micro,
  title={Micro-level estimation of poverty and inequality},
  author={Elbers, Chris and Lanjouw, Jean O and Lanjouw, Peter},
  journal={Econometrica},
  volume={71},
  number={1},
  pages={355--364},
  year={2003},
  publisher={JSTOR}
}

@article{cliff1970spatial,
  title={Spatial autocorrelation: a review of existing and new measures with applications},
  author={Cliff, Andrew D and Ord, Keith},
  journal={Economic Geography},
  volume={46},
  number={sup1},
  pages={269--292},
  year={1970},
  publisher={Taylor \& Francis}
}

@misc{gustafsson2024estimating,
  title={Estimating Aid Effectiveness in Fragile and Conflict-affected States: Evidence From Satellite-based Inference in Somalia},
  author={Gustafsson, Mikael P},
  year={2024}
}

@article{daoud_using_2023-1,
	title = {Using {Satellite} {Images} and {Deep} {Learning} to {Measure} {Health} and {Living} {Standards} in {India}},
	volume = {167},
	issn = {1573-0921},
	url = {https://doi.org/10.1007/s11205-023-03112-x},
	doi = {10.1007/s11205-023-03112-x},
	language = {en},
	number = {1},
	urldate = {2023-12-07},
	journal = {Social Indicators Research},
	author = {Daoud, Adel and Jordán, Felipe and Sharma, Makkunda and Johansson, Fredrik and Dubhashi, Devdatt and Paul, Sourabh and Banerjee, Subhashis},
	month = jun,
	year = {2023},
	pages = {475--505},
}

@article{daoud_statistical_2023,
	title = {Statistical {Modeling}: {The} {Three} {Cultures}},
	volume = {5},
	issn = {2644-2353, 2688-8513},
	shorttitle = {Statistical {Modeling}},
	url = {https://hdsr.mitpress.mit.edu/pub/uo4hjcx6/release/1},
	doi = {10.1162/99608f92.89f6fe66},
	language = {en},
	number = {1},
	urldate = {2023-01-28},
	journal = {Harvard Data Science Review},
	author = {Daoud, Adel and Dubhashi, Devdatt},
	month = jan,
	year = {2023},
}

@article{khan2022transformers,
  title={Transformers in vision: A survey},
  author={Khan, Salman and Naseer, Muzammal and Hayat, Munawar and Zamir, Syed Waqas and Khan, Fahad Shahbaz and Shah, Mubarak},
  journal={ACM computing surveys (CSUR)},
  volume={54},
  number={10s},
  pages={1--41},
  year={2022},
  publisher={ACM New York, NY}
}

@book{imbens2015causal,
  title={{Causal Inference in Statistics, Social, and Biomedical Sciences}},
  author={Imbens, Guido W and Rubin, Donald B},
  year={2015},
  publisher={Cambridge University Press}
}

@article{babenko2017poverty,
  title={Poverty mapping using convolutional neural networks trained on high and medium resolution satellite images, with an application in Mexico},
  author={Babenko, Boris and Hersh, Jonathan and Newhouse, David and Ramakrishnan, Anusha and Swartz, Tom},
  journal={arXiv preprint arXiv:1711.06323},
  year={2017}
}

@article{meng2014trio,
  title={A trio of inference problems that could win you a Nobel prize in statistics (if you help fund it)},
  author={Meng, Xiao-Li},
  journal={Past, Present, and Future of Statistical Science},
  pages={537--562},
  year={2014},
  publisher={CRC Press Boca Raton}
}

@article{suss2024geowealth,
  title={GEOWEALTH-US: Spatial wealth inequality data for the United States, 1960--2020},
  author={Suss, Joel and Kemeny, Tom and Connor, Dylan S},
  journal={Scientific Data},
  volume={11},
  number={1},
  pages={253},
  year={2024},
  publisher={Nature Publishing Group UK London}
}

@article{jerzak2024effect,
  title={Effect Heterogeneity with Earth Observation in Randomized Controlled Trials: Exploring the Role of Data, Model, and Evaluation Metric Choice},
  author={Jerzak, Connor T. and Vashistha, Ritwik and Daoud, Adel},
  journal={arXiv preprint arXiv:2407.11674},
  year={2024}
}

@article{burke2021using,
  title={Using satellite imagery to understand and promote sustainable development},
  author={Burke, Marshall and Driscoll, Anne and Lobell, David B and Ermon, Stefano},
  journal={Science},
  volume={371},
  number={6535},
  pages={eabe8628},
  year={2021},
  publisher={American Association for the Advancement of Science}
}

@inbook{Reddy_Daoud_2020, place={Cambridge}, title={Entitlements and Capabilities}, booktitle={The Cambridge Handbook of the Capability Approach}, publisher={Cambridge University Press}, author={Reddy, Sanjay G. and Daoud, Adel}, editor={Chiappero-Martinetti, Enrica and Osmani, Siddiqur and Qizilbash, MozaffarEditors}, year={2020}, pages={677–685}}

@article{Daoud_2017Malthus,
    author = {Daoud, Adel},
    title = "{Synthesizing the Malthusian and Senian approaches on scarcity: a realist account}",
    journal = {Cambridge Journal of Economics},
    volume = {42},
    number = {2},
    pages = {453-476},
    year = {2017},
    month = {05},
    issn = {0309-166X},
    doi = {10.1093/cje/bew071},
    url = {https://doi.org/10.1093/cje/bew071},
    eprint = {https://academic.oup.com/cje/article-pdf/42/2/453/24145834/bew071.pdf},
}

@article{yeh2020using,
  title={Using publicly available satellite imagery and deep learning to understand economic well-being in Africa},
  author={Yeh, Christopher and Perez, Anthony and Driscoll, Anne and Azzari, George and Tang, Zhongyi and Lobell, David and Ermon, Stefano and Burke, Marshall},
  journal={Nature Communications},
  volume={11},
  number={1},
  pages={2583},
  year={2020},
  publisher={Nature Publishing Group UK London}
}

@incollection{kakooei2022remote,
  title={Remote sensing technology for postdisaster building damage assessment},
  author={Kakooei, Mohammad and Ghorbanian, Arsalan and Baleghi, Yasser and Amani, Meisam and Nascetti, Andrea},
  booktitle={Computers in Earth and Environmental Sciences},
  pages={509--521},
  year={2022},
  publisher={Elsevier}
}

@inproceedings{xie2016transfer,
  title={Transfer learning from deep features for remote sensing and poverty mapping},
  author={Xie, Michael and Jean, Neal and Burke, Marshall and Lobell, David and Ermon, Stefano},
  booktitle={Proceedings of the AAAI Conference on Artificial Intelligence},
  volume={30},
  number={1},
  year={2016}
}

@book{world2018mainstreaming,
  title={Mainstreaming the use of remote sensing data and applications in operational contexts},
  author={{World Bank Group}},
  year={2018},
  publisher={World Bank}
}

@techreport{cerda2024causal,
  title={Causal evaluation of humanitarian aid on food security},
  author={Cerd{\`a}-Bautista, Jordi and T{\'a}rraga, Jos{\'e} Mar{\'\i}a and Sitokonstantinou, Vasileios and Camps-Valls, Gustau},
  year={2024},
  institution={Copernicus Meetings}
}

@article{vaswani2017attention,
  title={Attention is all you need},
  author={Vaswani, A},
  journal={Advances in Neural Information Processing Systems},
  year={2017}
}

@book{prince2023understanding,
        author = "Simon J.D. Prince",
        title = "Understanding Deep Learning",
        publisher = "The MIT Press",
        year = 2023,
        url = "http://udlbook.com"
}

@inproceedings{gordon2023remote,
  title={Remote control: Debiasing remote sensing predictions for causal inference},
  author={Gordon, Matthew and Ayers, Megan and Stone, Eliana and Sanford, Luke},
  booktitle={Proc. Int. Conf. Learn. Represent. Workshops},
  number={22},
  year={2023}
}

@article{rolf2021generalizable,
  title={{A Generalizable and Accessible Approach to Machine Learning with Global Satellite Imagery}},
  author={Rolf, Esther and Proctor, Jonathan and Carleton, Tamma and Bolliger, Ian and Shankar, Vaishaal and Ishihara, Miyabi and Recht, Benjamin and Hsiang, Solomon},
  journal={Nature Communications},
  volume={12},
  number={1},
  pages={4392},
  year={2021},
  publisher={Nature Publishing Group UK London}
}

@article{angelopoulos2023prediction,
  title={Prediction-powered inference},
  author={Angelopoulos, Anastasios N and Bates, Stephen and Fannjiang, Clara and Jordan, Michael I and Zrnic, Tijana},
  journal={Science},
  volume={382},
  number={6671},
  pages={669--674},
  year={2023},
  publisher={American Association for the Advancement of Science}
}

@article{angelopoulos2021gentle,
  title={A gentle introduction to conformal prediction and distribution-free uncertainty quantification},
  author={Angelopoulos, Anastasios N and Bates, Stephen},
  journal={arXiv preprint arXiv:2107.07511},
  year={2021}
}

@article{kakooei2024increasing,
  title={Increasing the confidence of predictive uncertainty: earth observations and deep learning for poverty estimation},
  author={Kakooei, Mohammad and Daoud, Adel},
  journal={IEEE Transactions on Geoscience and Remote Sensing},
  year={2024},
  publisher={IEEE}
}

@article{olofsson2014good,
  title={Good practices for estimating area and assessing accuracy of land change},
  author={Olofsson, Pontus and Foody, Giles M and Herold, Martin and Stehman, Stephen V and Woodcock, Curtis E and Wulder, Michael A},
  journal={Remote sensing of Environment},
  volume={148},
  pages={42--57},
  year={2014},
  publisher={Elsevier}
}

@techreport{aiken2021machine,
  title={Machine learning and mobile phone data can improve the targeting of humanitarian assistance},
  author={Aiken, Emily and Bellue, Suzanne and Karlan, Dean and Udry, Christopher R and Blumenstock, Joshua},
  year={2021},
  institution={National Bureau of Economic Research}
}

@article{perkins2009satellite,
  title={Satellite imagery and the spectacle of secret spaces},
  author={Perkins, Chris and Dodge, Martin},
  journal={Geoforum},
  volume={40},
  number={4},
  pages={546--560},
  year={2009},
  publisher={Elsevier}
}

@article{ogburn2024causal,
  title={Causal inference for social network data},
  author={Ogburn, Elizabeth L and Sofrygin, Oleg and Diaz, Ivan and Van der Laan, Mark J},
  journal={Journal of the American Statistical Association},
  volume={119},
  number={545},
  pages={597--611},
  year={2024},
  publisher={Taylor \& Francis}
}

@inproceedings{wang2024skyscript,
  title={Skyscript: A large and semantically diverse vision-language dataset for remote sensing},
  author={Wang, Zhecheng and Prabha, Rajanie and Huang, Tianyuan and Wu, Jiajun and Rajagopal, Ram},
  booktitle={Proceedings of the AAAI Conference on Artificial Intelligence},
  volume={38},
  number={6},
  pages={5805--5813},
  year={2024}
}

@article{dionelis2024evaluating,
  title={Evaluating and Benchmarking Foundation Models for Earth Observation and Geospatial AI},
  author={Dionelis, Nikolaos and Fibaek, Casper and Camilleri, Luke and Luyts, Andreas and Bosmans, Jente and Saux, Bertrand Le},
  journal={arXiv preprint arXiv:2406.18295},
  year={2024}
}

@article{scholkopf2021toward,
  title={Toward causal representation learning},
  author={Sch{\"o}lkopf, Bernhard and Locatello, Francesco and Bauer, Stefan and Ke, Nan Rosemary and Kalchbrenner, Nal and Goyal, Anirudh and Bengio, Yoshua},
  journal={Proceedings of the IEEE},
  volume={109},
  number={5},
  pages={612--634},
  year={2021},
  publisher={IEEE}
}

@book{pearl2009causality,
  title={Causality: Models, Reasoning, and Inference},
  author={Pearl, Judea},
  year={2009},
  publisher={Cambridge University Press}
}

@article{orshansky1965counting,
  title={Counting the poor: Another look at the poverty profile},
  author={Orshansky, Mollie},
  journal={Soc. Sec. Bull.},
  volume={28},
  pages={3},
  year={1965},
  publisher={HeinOnline}
}

@book{oecd2024development,
  title={Development Co-operation report 2024: tackling poverty and Inequalities through the green transition},
  author={OECD.},
  year={2024},
  publisher={OECD Publications Centre}
}

@article{beck2007finance,
  title={Finance, inequality and the poor},
  author={Beck, Thorsten and Demirg{\"u}{\c{c}}-Kunt, Asli and Levine, Ross},
  journal={Journal of economic growth},
  volume={12},
  pages={27--49},
  year={2007},
  publisher={Springer}
}

@article{ridley2020poverty,
  title={Poverty, depression, and anxiety: Causal evidence and mechanisms},
  author={Ridley, Matthew and Rao, Gautam and Schilbach, Frank and Patel, Vikram},
  journal={Science},
  volume={370},
  number={6522},
  pages={eaay0214},
  year={2020},
  publisher={American Association for the Advancement of Science}
}

@incollection{ribot2017cause,
  title={Cause and response: vulnerability and climate in the Anthropocene},
  author={Ribot, Jesse},
  booktitle={New Directions in Agrarian Political Economy},
  pages={27--66},
  year={2017},
  publisher={Routledge}
}

@article{molina2014small,
  title={Small area estimation of general parameters with application to poverty indicators: A hierarchical Bayes approach},
  author={Molina, Isabel and Nandram, Balgobin and Rao, JNK},
  year={2014}
}

@article{keele2015geographic,
  title={Geographic boundaries as regression discontinuities},
  author={Keele, Luke J and Titiunik, Rocio},
  journal={Political Analysis},
  volume={23},
  number={1},
  pages={127--155},
  year={2015},
  publisher={Cambridge University Press}
}

@article{butts2021difference,
  title={Difference-in-differences estimation with spatial spillovers},
  author={Butts, Kyle},
  journal={arXiv preprint arXiv:2105.03737},
  year={2021}
}

@article{dube2014spatial,
  title={A spatial difference-in-differences estimator to evaluate the effect of change in public mass transit systems on house prices},
  author={Dub{\'e}, Jean and Legros, Di{\`e}go and Th{\'e}riault, Marius and Des Rosiers, Fran{\c{c}}ois},
  journal={Transportation Research Part B: Methodological},
  volume={64},
  pages={24--40},
  year={2014},
  publisher={Elsevier}
}

@article{betz2020spatial,
  title={Spatial interdependence and instrumental variable models},
  author={Betz, Timm and Cook, Scott J and Hollenbach, Florian M},
  journal={Political science research and methods},
  volume={8},
  number={4},
  pages={646--661},
  year={2020},
  publisher={Cambridge University Press}
}

@incollection{kelejian2004instrumental,
  title={Instrumental variable estimation of a spatial autoregressive model with autoregressive disturbances: Large and small sample results},
  author={Kelejian, Harry H and Prucha, Ingmar R and Yuzefovich, Yevgeny},
  booktitle={Spatial and spatiotemporal econometrics},
  volume={18},
  pages={163--198},
  year={2004},
  publisher={Emerald Group Publishing Limited}
}

@article{nilsson2021directed,
  title={A Directed Acyclic Graph for Interactions},
  author={Nilsson, Anton and Bonander, Carl and Str{\"o}mberg, Ulf and Bj{\"o}rk, Jonas},
  journal={International journal of epidemiology},
  volume={50},
  number={2},
  pages={613--619},
  year={2021},
  publisher={Oxford University Press}
}

@article{schwartz2011transportability,
  title={Transportability and causal generalization},
  author={Schwartz, Sharon and Gatto, Nicolle M and Campbell, Ulka B},
  journal={Epidemiology},
  volume={22},
  number={5},
  pages={745--746},
  year={2011},
  publisher={LWW}
}

@article{alkire2011counting,
  title={Counting and multidimensional poverty measurement},
  author={Alkire, Sabina and Foster, James},
  journal={Journal of public economics},
  volume={95},
  number={7-8},
  pages={476--487},
  year={2011},
  publisher={Elsevier}
}

@inproceedings{pearl2011transportability,
  title={Transportability of causal and statistical relations: A formal approach},
  author={Pearl, Judea and Bareinboim, Elias},
  booktitle={Proceedings of the AAAI Conference on Artificial Intelligence},
  volume={25},
  number={1},
  pages={247--254},
  year={2011}
}

@article{kakooei2024analyzing,
  title={Analyzing Poverty through Intra-Annual Time-Series: A Wavelet Transform Approach},
  author={Kakooei, Mohammad and Solska, Klaudia and Daoud, Adel},
  journal={arXiv preprint arXiv:2411.02855},
  year={2024}
}

@article{runge2019inferring,
  title={Inferring causation from time series in Earth system sciences},
  author={Runge, Jakob and Bathiany, Sebastian and Bollt, Erik and Camps-Valls, Gustau and Coumou, Dim and Deyle, Ethan and Glymour, Clark and Kretschmer, Marlene and Mahecha, Miguel D and Mu{\~n}oz-Mar{\'\i}, Jordi and others},
  journal={Nature communications},
  volume={10},
  number={1},
  pages={2553},
  year={2019},
  publisher={Nature Publishing Group UK London}
}

\section*{Appendix}

The search terms used were: `( `eo' OR `earth observation' OR `satellite image*' OR `remote sensing' OR `earth science' OR `earth’s system' OR `environmental eco*' OR `environmental science*' OR `geosci*') AND (`machine learning' OR `ml' OR `ai' OR `deep learning' ) AND ( `causal inf*' OR `causal relation*' OR `causality' OR `causal learning' OR `causal effect*' OR `policy eval*' OR `causal impact' OR `causal discovery' OR `treatment effect*' OR `causal representation')''. We limited the scope of the survey to between 2011 and 2024 (as few papers appear before 2011). This resulted in 101 papers and 41 preprints.

We analyzed the SCOPUS articles and pre-prints in the following way. First, bigrams from the abstracts were analyzed; Table \ref{tab:Bigrams} shows bigrams that appeared over 12 times.  Beyond the expected results, some of the notable bigrams are time series, which occurred 38 times, earth system/ sciences 28 and 13, and causal discovery at 22 times. There are some trends in the type of data being processed, like time series or remotely sensed variety, the domains being studied, like earth sciences, and the types of methods or concepts, like causal discovery and Granger causality, that repeatedly show up in the nascent literature. 

\begin{table}[H]
\centering
\begin{tabular}{rlr}
    \hline \hline
 & {\bf Abstract Bigram} & {\bf Count}\\ 
  \hline
1 & machine learning &  99 \\ 
  2 & remote sensing &  65 \\ 
  3 & deep learning &  60 \\ 
  4 & time series &  38 \\ 
  5 & causal inference &  34 \\ 
  6 & earth system &  28 \\ 
  7 & causal relationships &  24 \\ 
  8 & data driven &  24 \\ 
  9 & artificial intelligence &  22 \\ 
  10 & causal discovery &  22 \\ 
  11 & neural network &  21 \\ 
  12 & causal relationship &  15 \\ 
  13 & granger causality &  15 \\ 
  14 & real world &  15 \\ 
  15 & satellite imagery &  14 \\ 
  16 & earth sciences &  13 \\ 
  17 & treatment effect &  12 \\ 
   \hline
\end{tabular}
\caption{Top bigrams among papers found in the literature search.}\label{tab:Bigrams}
\end{table}

We then evaluate where papers are being published, Table \ref{tab:JournalThemes} shows the various journal themes, with remote sensing journals having the greatest number followed by earth and environment, and then by AI/ML/CV publications. This goes to show that technologically minded journals have been utilized, followed by earth and environmental science-related ones. Next, the content of the paper was evaluated by themes, and surprisingly, most of the papers were grounded in earth and environmental sciences, while only 21 of the papers focused only on technical topics. For the geography of poverty, we find that there are few papers dealing with economic topics, being far eclipsed by natural science topics. 

Another finding is that technical development is being outpaced by applications. The majority of papers focus on using machine learning and causal inference methods to study substantive questions in earth and environmental sciences. Comparatively, fewer papers are solely dedicated to extending or developing novel methodologies at the intersection of these fields. This suggests there may be opportunities for more dedicated technical and methodological research to expand the toolkit for applying ML and causal inference to geospatial data.

\begin{table}[H]
\centering
\begin{tabular}{rlr}
  \hline \hline
 & {\bf Journal Theme} & {\bf Count}\\ 
  \hline
1 & Remote sensing &  18 \\ 
  2 & Earth and environment &  14 \\ 
  3 & AI ML CV&   10\\ 
  4 & Economics &   3 \\ 
  5& Public health &   1 \\ 
   \hline
\end{tabular}
\caption{Journal themes in quantitative review of EO-ML literature.}\label{tab:JournalThemes}
\end{table}

In the literature found, four papers used double machine learning, five chose Granger causality, and nine used some network or graphical approach. Many of the outcomes were focused on environmental factors like vegetation, atmospheric measurements, or hydrological features. Only 12 papers were focused on socio-economic outcomes such as IWI or local development, e.g., from infrastructure projects like roads, airports, and the electric grid.  

\begin{table}[H]
\centering
\begin{tabular}{rlr}
  \hline \hline
 & {\bf Paper Theme} & {\bf Count}\\ 
  \hline
1 & Earth and environment &  41 \\ 
  2 & AI ML CV &  21 \\ 
  3 & Economics &   9 \\ 
  4 & Public health &   1 \\ 
   \hline
\end{tabular}
\caption{Paper themes in quantitative review of EO-ML literature.}\label{tab:PaperThemes}
\end{table}

Taken together, this review suggests that the intersection of causal inference, machine learning, and earth observation is a small but growing field.

\end{document}